\documentclass[10pt,twocolumn,letterpaper]{article}

\usepackage{iccv}
\usepackage{times}
\usepackage{epsfig}
\usepackage{graphicx}
\usepackage{amsmath}
\usepackage{amssymb}

\usepackage{algorithm}
\usepackage{algorithmic}
\usepackage{listings} 

\usepackage{bm}
\usepackage{subcaption}
\usepackage{bbm}
\usepackage{multicol}
\usepackage{multirow}
\usepackage{color}
\usepackage{caption}
\usepackage{hhline}
\usepackage{pifont}
\usepackage{threeparttable}
\usepackage{makecell}
\usepackage{wrapfig}
\usepackage{colortbl}  
\usepackage{url}
\usepackage{enumitem}
\usepackage{tablefootnote}
\usepackage{booktabs}
\usepackage{diagbox}
\usepackage[table]{xcolor}
\usepackage[title]{appendix}

\usepackage{tikz}
\usepackage{pgfplots}
\usepackage{pgfplotstable}

\newlength\savewidth\newcommand\shline{\noalign{\global\savewidth\arrayrulewidth
  \global\arrayrulewidth 1pt}\hline\noalign{\global\arrayrulewidth\savewidth}}
  
\newcolumntype{x}[1]{>{\centering\arraybackslash}p{#1pt}}
\newcolumntype{y}[1]{>{\raggedright\arraybackslash}p{#1pt}}
\newcolumntype{z}[1]{>{\raggedleft\arraybackslash}p{#1pt}}
\renewcommand{\paragraph}[1]{\vspace{1.25mm}\noindent\textbf{#1}}
\definecolor{deemph}{gray}{0.6}

\definecolor{neucolor}{RGB}{160,160,160}
\definecolor{bblue}{rgb}{0.855,0.933,0.98}

\newcommand{\graybox}[1]{\colorbox{lightgray!28}{#1}}

\newcommand{\myparagraph}[1]{{\vspace{.4em} \noindent \bf #1}}

\newcommand{\xmark}{\ding{55}}
\newcommand{\cmark}{\ding{51}}

\newcommand{\mj}{$\mathcal{J}$}
\newcommand{\mf}{$\mathcal{F}$}
\newcommand{\mjf}{$\mathcal{J}\&\mathcal{F}$}

\newcommand{\mjs}{$\mathcal{J}_s$}
\newcommand{\mfs}{$\mathcal{F}_s$}
\newcommand{\mju}{$\mathcal{J}_u$}
\newcommand{\mfu}{$\mathcal{F}_u$}
\newcommand{\mg}{$\mathcal{G}$}

\definecolor{codeblue}{rgb}{0.25,0.5,0.5}
\definecolor{codegreen}{rgb}{0,0.6,0}
\definecolor{codekw}{rgb}{0.85, 0.18, 0.50}
\usepackage{etoolbox}
\makeatletter
\AfterEndEnvironment{algorithm}{\let\@algcomment\relax}
\AtEndEnvironment{algorithm}{\kern2pt\hrule\relax\vskip3pt\@algcomment}
\let\@algcomment\relax
\newcommand\algcomment[1]{\def\@algcomment{\footnotesize#1}}
\renewcommand\fs@ruled{\def\@fs@cfont{\bfseries}\let\@fs@capt\floatc@ruled
  \def\@fs@pre{\hrule height.8pt depth0pt \kern2pt}%
  \def\@fs@post{}%
  \def\@fs@mid{\kern2pt\hrule\kern2pt}%
  \let\@fs@iftopcapt\iftrue}
\makeatother

\definecolor{redstar}{RGB}{240,47,29}
\definecolor{curly}{RGB}{187,101,183}
\definecolor{lightsalmon}{RGB}{255,160,122}
\definecolor{limegreen}{RGB}{50,205,50}
\definecolor{grey}{RGB}{190,190,190}
\definecolor{lightsteelblue}{RGB}{202,225,255}

\definecolor{poscolor}{RGB}{83,161,81}
\definecolor{negcolor}{RGB}{103,81,165}
\definecolor{neucolor}{RGB}{190,190,190}

\def\modelname{UniRef\xspace}
\def\modelnameplus{UniRef++\xspace}

\definecolor{linkColor}{rgb}{0.18,0.39,0.62}
\usepackage[pagebackref=true,breaklinks=true,colorlinks,citecolor=linkColor]{hyperref}

\iccvfinalcopy 

\def\iccvPaperID{xxxx} 

\begin{document}

\title{UniRef++: Segment Every Reference Object in Spatial and Temporal Spaces}

\author
{
Jiannan Wu$^{1}$, 
~~~
Yi Jiang$^{2}$,
~~~
Bin Yan$^{3}$, 
~~~
Huchuan Lu$^{3}$, 
~~~
Zehuan Yuan$^{2}$, 
~~~
Ping Luo$^{1,4}$
\\[0.2cm]
${^1}$The University of Hong Kong ~~~
${^2}$ByteDance ~~~ \\[0.1cm]
${^3}$Dalian University of Technology ~~~
${^4}$Shanghai AI Laboratory 
}

\maketitle

\begin{abstract}

   The reference-based object segmentation tasks, namely referring image segmentation (RIS), few-shot image segmentation (FSS), referring video object segmentation (RVOS), and video object segmentation (VOS), aim to segment a specific object by utilizing either language or annotated masks as references. Despite significant progress in each respective field, current methods are task-specifically designed and developed in different directions, which hinders the activation of multi-task capabilities for these tasks. In this work, we end the current fragmented situation and propose \modelnameplus to unify the four reference-based object segmentation tasks with a single architecture. At the heart of our approach is the proposed UniFusion module which performs multiway-fusion for handling different tasks with respect to their specified references. And a unified Transformer architecture is then adopted for achieving instance-level segmentation. With the unified designs, \modelnameplus can be jointly trained on a broad range of benchmarks and can flexibly complete multiple tasks at run-time by specifying the corresponding references. We evaluate our unified models on various benchmarks. Extensive experimental results indicate that our proposed \modelnameplus achieves state-of-the-art performance on RIS and RVOS, and performs competitively on FSS and VOS with a parameter-shared network. Moreover, we showcase that the proposed UniFusion module could be easily incorporated into the current advanced foundation model SAM and obtain satisfactory results with parameter-efficient finetuning. Codes and models are available at \url{https://github.com/FoundationVision/UniRef}.

\end{abstract}

\vspace{-1mm}
\section{Introduction} \label{intro}
\vspace{-1mm}

The reference-guided object segmentation aims at segmenting the specified object with the given references (\emph{e.g.}, language or annotated mask). The four representative tasks include referring image segmentation (RIS)~\cite{yu2016refcoco}, few-shot segmentation(FSS)~\cite{ravi2016optimization}, referring video object segmentation (RVOS)~\cite{khoreva2019video} and semi-supervised video object segmentation (VOS)~\cite{pont2017davis}, which are the fundamental tasks for vision understanding. Over time, many advanced methods have ballooned in their respective fields and rapidly improves the state-of-the-art performance.

Despite witnessing the significant progress, these tasks are separately tackled with specialized designed models. In that regard, the individual methods need extra training time and produce different sets of model weights on each task. This would cause expensive computational cost and yield redundant parameters. More importantly, the independent designs prevent the synergy and facilitation of different tasks. We argue that the current fragmented situation is unnecessary as the four tasks have essentially the same definition in a high-level aspect: they all use the references (language or annotated mask) as guidance to perform the per-pixel segmentation of the target object. This motivates us to build a unified model within the same parameters, which can perform different tasks at run-time by specifying the corresponding references.

\begin{figure}[t]
\begin{center}
\includegraphics[width=0.48\textwidth]{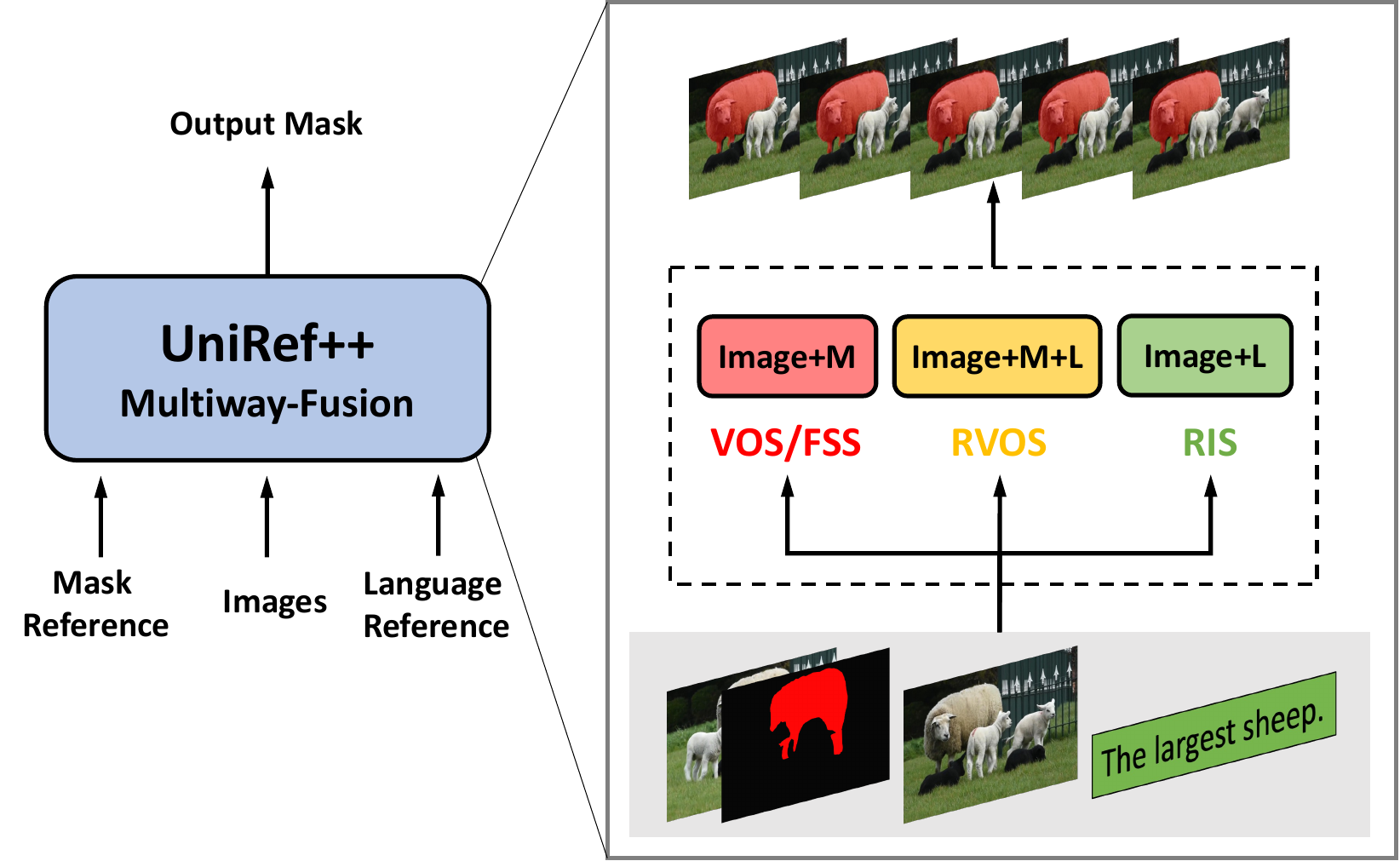}
\end{center}
\vspace{-2mm}
\caption{A single, jointly trained \modelnameplus can perform three different reference-based tasks by specifying the corresponding references. `L' and `M' represent language and mask references, respectively.}
\label{fig:teaser}
\vspace{-5mm}
\end{figure}

Towards the unification of reference-based object segmentation tasks, it poses great challenges in connecting the isolated landscapes as a whole: 
\textbf{(1) The mainstream methods in different fields vary greatly.} RIS methods~\cite{ye2019cmsa, huang2020cmpc, yang2022lavt} mostly focus on the deep cross-modal fusion of vision and language information. FSS community is advocating the correlation-based methods~\cite{hong2022vat, xiong2022dacm} for dense semantic correspondence. VOS has been long dominated by the space-time memory network~\cite{oh2019stm, cheng2021stcn, yang2021aot, cheng2022xmem} for pixel-level matching. While the recenet RVOS methods heavily rely on the query-based methods~\cite{botach2022mttr, wu2022referformer}.
\textbf{(2) The image-level methods cannot be simply extended to the video domain.} The image tasks only require to segment the referred target in a single image. For the video tasks, however, the objects may encounter occlusion, fast motion or disapperance-reappearance in many complex scenes, which requires the networks to leverage the spatio-temporal information to track the objects throughout the whole video. Hence, simply adopting the image-level methods for each frame independently cannot ensure the temporal consistency for target object in videos.
\textbf{(3) The video tasks (VOS and RVOS) are solved in two different paradigms currently.} The previous state-of-the-art RVOS methods~\cite{botach2022mttr, wu2022referformer} take the whole video as input and generate the prediction results for all frames in one single step, which termed as offline methods. In contrast, VOS methods~\cite{oh2019stm, cheng2021stcn} operate in an online fashion where they readout the historical features to propagate the target masks frame by frame.

In this work, we conquer the challenges above and propose a unified model, \modelnameplus, for the reference-based object segmentation tasks. The key idea behind our approach is to formulate all four tasks as instance-level segmentation problems, and the information of references can be injected into the network through an attention-based fusion process regardless of their modalities. As illustrated in Figure~\ref{fig:teaser}, for different tasks, \modelnameplus receives the current frame and utilizes the corresponding references to perform the fusion process, termed as multiway-fusion. Specifically, the annotated mask for reference image is leveraged as reference for FSS and VOS. The reference comes from language description for RIS. And we emphasize that, for RVOS, both the language and mask references are used. This design not only tackles RVOS in an online fashion, but also can utilize the historical information for mask propagation to ensure the temporal consistency for target object, establishing a new paradigm for RVOS. 

Practically, we introduce a UniFusion module to fuse the visual features and the specified references. Afterwards, the visual features of current frame are fed into a unified Transformer architecture, where queries are employed for instance-level segmentation of the target object. Thanks to the unified architecture, our model can be jointly trained on the broad range of benchmarks to learn the general knowledge, and can flexibly perform multi-tasks at run-time by specifying the corresponding references. 

To summarize, our contributions are as follows:

\vspace{-2mm}
\begin{itemize}[leftmargin=*]
\item We propose \modelnameplus, a unified model to perform four reference-based object segmentation tasks (RIS, FSS, RVOS, VOS) with the same model weights.
\vspace{-2mm}
\item We introduce a UniFusion module to inject the reference information into the network regardless of their modalities. And we establish a new online paradigm for RVOS by leveraging both language and mask as references.
\vspace{-2mm}
\item Extensive experiments demonstrate that our models achieve state-of-the-art performance for RIS and RVOS, and perform competitively for FSS, VOS.
\end{itemize}

\vspace{-1mm}
\section{Related Work} \label{related_work}
\vspace{-1mm}

\subsection{Unified Model}

Towards achieving general artificial intelligence, the vision community has clearly witnessed the trend of building unified models recently. One line of works~\cite{alayrac2022flamingo, yuan2021florence, yu2022coca, wang2022beitv3, chen2021pix2seq, chen2022pixel2seqv2, zhu2022uni-perceiver, kolesnikov2022uvim, wang2022ofa, lu2022unified-io} is to design the \emph{general interface} for vision or vision-language (VL) tasks. For example, Unified-IO~\cite{lu2022unified-io} unifies broad range of image-level tasks (\emph{e.g.}, image classification~\cite{deng2009imagenet}, image caption~\cite{chen2015coco-captions} and VQA~\cite{antol2015vqa}) in a sequence-to-sequence generation paradigm. Another line of works~\cite{cheng2022mask2former, kamath2021mdetr, li2022glip, zhang2022glipv2, yan2022unicorn, zou2022x-decoder, zhang2023openseed, liu2023groundingdino, jain2022oneformer, bhattacharjee2022mult, athar2023tarvis, yan2023uninext, wu2023glee} is to build the \emph{unified architecture} for the closely related tasks. GLIP~\cite{li2022glip} formulates both the object detection and phrase grounding tasks as the word-region alignment problem. OneFormer~\cite{jain2022oneformer} rules the universal image segmentation tasks with a single Transformer network. Unicorn~\cite{yan2022unicorn} proposes the designs of target priors to tackle four tracking tasks. However, these methods with unified architecture either focus on the image domain or only consider the visual-only tasks. We aim to bridge the gap by building the unified model for the reference-based object segmentation tasks.

\begin{figure*}[t]
\begin{center}
\includegraphics[width=0.98\textwidth]{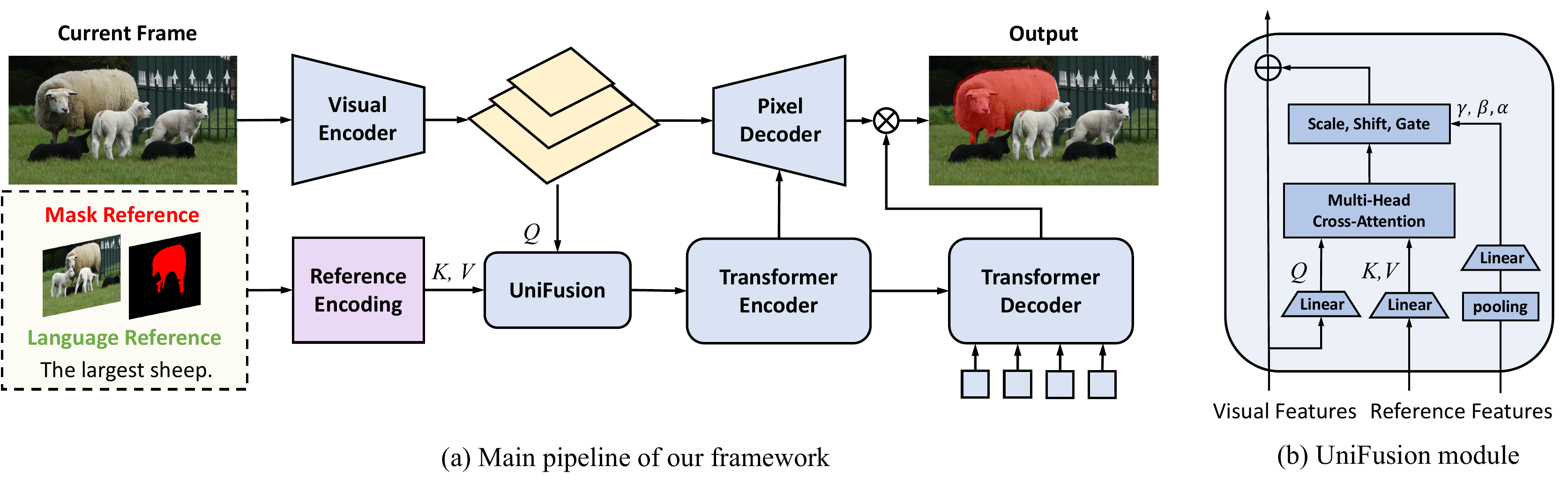}
\end{center}
\vspace{-6mm}
\caption{Illustration of (a) the overall framework of \modelnameplus. For sake of clarity, we omit the visualization of prediction heads which are on top of Transformer decoder. The core network (in \textcolor{lightsteelblue}{\textbf{blue}}) is shared for all tasks. (b) The details of UniFusion module. The reference features come from the language or mask references.}
\label{fig:network}
\vspace{-3mm}
\end{figure*}


\vspace{-1mm}
\subsection{Task-specific Object Segmentation} 
\vspace{-1mm}

\myparagraph{\textit{Referring Image Segmentation}.} The objective of RIS~\cite{hu2016segmentation} is to generate a pixel-level mask for the target object described by a language description in an image. Prior research has primarily focused on the multi-modal feature interaction techniques, either employing attention mechanism in CNNs~\cite{yu2018mattnet, chen2019see, ye2019cmsa, huang2020cmpc, hu2020brinet, jing2021locate, yang2022lavt} or using multi-modal Transformers~\cite{ding2021vlt, kim2022restr, wang2022cris}. As RIS is closely related to referring expression comprehension (REC), which aims to predict the bounding box of the referred object, some works~\cite{luo2020mcn, li2021reftr, zhu2022seqtr} also explore the unified frameworks that can accomplish these two tasks simultaneously.

\myparagraph{\textit{Few-shot Segmentation}.} FSS task aims to provide a mask prediction for a query image with a handful of support samples. The early methods mainly utilize the prototype-based networks~\cite{wang2019panet, yang2020prototype, liu2020part}, where the network computes the class prototype by a masked average pooling from support set and then refine the query image features with the prototype information. As these methods have significant information loss due to the pooling operation, the correlation-based methods~\cite{hong2022vat, xiong2022dacm, hu2019attention, wang2020dan} have been proposed to model the densely pixel-level relationships between query and support images.

\myparagraph{\textit{Referring Video Object Segmentation}.} RVOS can be considered as an extension of RIS in the video domain. Some previous methods process the video frames independently~\cite{ye2019cmsa, zhao2022mvtt, ding2022lbdt, li2022yofo} or simply adopt 3D CNNs~\cite{wang2019asymmetric, mcintosh2020visual, wang2020context} to extract the spatio-temporal features for a video clip. Recently, state-of-the-art methods~\cite{botach2022mttr, wu2022referformer} are based on query-based Transformers and process the videos in an offline fashion. They receive the whole video as input and employ the queries to segment and track the target object simultaneously. However, such methods are not suitable for long videos or ongoing videos. In contrast to these works, our \modelnameplus belongs to the online method and can utilize the historical information for mask propagation, which can ensure the temporal consistency of target object and improve segmentation accuracy.

\myparagraph{\textit{Video Object Segmentation}.} Given a video with the target mask annotations in the first frame, the VOS algorithms need to propagate the masks to the entire video. The previous approaches could be broadly categorized into two groups: (i) Template-based methods. These works~\cite{voigtlaender2019feelvos, wang2019siammask, voigtlaender2020siam-rcnn, robinson2020frtm, chen2020sat} regard the annotated frame as template and investigate how to fuse the template information into the current frame.  (ii) Memory-based methods. The pioneering work STM~\cite{oh2019stm} leverages a memory network to embed past-frame predictions and learns the space-time pixel-level correspondence on the memory to propagate the mask information. This type of works has achieved significant improvement and dominated the VOS community. The subsequent works mainly focus on improving memorized embeddings~\cite{yang2020cfbi, yang2021cfbi+, li2022rde, liang2020afb-ur, xu2022rpcm, mao2021joint}, designing novel memory networks~\cite{xie2021rmnet, cheng2022xmem, kei2023cutie} or proposing reliable memory readout strategies~\cite{seong2021hmmn, cheng2021stcn, yang2021aot, yang2022deaot}. These previous works view the VOS task as the pixel-level binary classification problem, lacking the understanding of object. Different from theirs, we tackle the VOS as the instance segmentation problem.
\vspace{-1mm}
\section{Method} \label{sec:method}
\vspace{-1mm}

\subsection{Overview}

We present \modelnameplus, a simple and unified architecture that can segment arbitrary objects with the given references in images and videos. Conceptually, it allows us to train a single network on all related benchmarks and simultaneously solves the aforementioned tasks (RIS, FSS, RVOS, VOS).

The overall architecture of \modelnameplus is illustrated in Fig.~\ref{fig:network}\textcolor{red}{a}. Our framework consists of a visual encoder, two reference encoders (for text and mask, respectively),  a proposed UniFusion module and a transformer-based object detector. Given an image $\bm{I} \in \mathbb{R}^{H \times W \times 3}$ and the corresponding references, we first use the visual encoder $\textbf{Enc}_{V}$ to extract the multi-scale features $\bm{\mathcal{F}} = \left \{ \bm{F}_{\ell} \right \}_{\ell =1}^{4}$ of current image, where $\ell$ denotes the level index of the hierarchical visual features, with the spatial strides from 4 to 32. Then the reference encoders are applied to encode the reference information, followed by the UniFusion module to inject the information into the visual features. Finally, the network can produce a binary mask for the target object $\bm{m} \in \mathbb{R}^{H \times W}$ via a unified transformer-based architecture. 

In the following subsections, we are going to details of \modelnameplus by introducing the reference encoding (Sec.~\ref{sec:referennce}), the multi-scale UniFusion module (Sec.~\ref{sec:fusion}), a unified encoder-decoder architecture (Sec.~\ref{sec:arch}), the training and inference process of \modelnameplus (Sec.~\ref{sec:train}). 

\subsection{Reference Encoding} \label{sec:referennce}

In this part, we introduce how to encode the reference information for the four reference-based tasks. Before that, we would like to clarify that the only task-specific design of \modelnameplus is to use different reference encoders (text encoder $\textbf{Enc}_{T}$ and mask encoder $\textbf{Enc}_{M}$) for processing different modalities.

\myparagraph{\textit{Few-shot Segmentation and Video Object Segmentation}.} For FSS and VOS tasks, the mask annotation for the reference image is provided as the reference. And the network needs to propagate the mask throughout the video. Inspired by the spirit of STCN~\cite{cheng2021stcn} that computing the similarity of two frames for once, we use the same visual encoder $\textbf{Enc}_{V}$ to extract the hierarchical visual features $\bm{\mathcal{F}_{V}^{\rm f}} = \left \{ \bm{F}_{V,\ell}^{\rm f} \right \}$ of reference frame $\bm{I}_\text{ref}$. Then, a lightweight mask encoder (\emph{e.g.}, ResNet18~\cite{he2016resnet}) receives the reference frame $\bm{I}_\text{ref}$, object mask annotation $\bm{m}_{o}$ and the encoded frame features $\bm{\mathcal{F}_{V}^{\rm f}}$ to generate the multi-scale mask features $\bm{\mathcal{F}_{V}^{\rm m}} = \left \{ \bm{F}_{V,\ell}^{\rm m} \right \}$ for the target object in reference frame. Here, $\ell = 2,3,4$ for $\bm{F}_{V,\ell}^{\rm f}$ and $\bm{F}_{V,\ell}^{\rm m}$.

\vspace{-2mm}

\begin{equation}
    \bm{\mathcal{F}}_{V}^{f} = \textbf{Enc}_{V}(\bm{I}_\text{ref})
    \label{eq:enc_Vref}
\end{equation}

\vspace{-5mm}
\begin{equation}
    \bm{\mathcal{F}}_{V}^{m} = \textbf{Enc}_{M}(\bm{I}_\text{ref}, \bm{m}_{o}, \bm{\mathcal{F}}_{V}^{f})
    \label{eq:enc_Vmask}
\end{equation}

\vspace{-2mm}

\myparagraph{\textit{Referring Image Segmentation}.} The reference for RIS task is the language description $\bm{T}$. To encode such linguistic information, we apply an off-the-shelf text encoder (\textit{e.g.}, BERT~\cite{devlin2018bert} or RoBERTa~\cite{liu2019roberta}) to extract the language features $\bm{F}_{T} \in \mathbb{R}^{L \times C}$, where $L$ is the sentence length and $C$ is the channel dimension. 

\vspace{-3mm}

\begin{equation}
    \bm{F}_{T} = \textbf{Enc}_{T}(\bm{T})
    \label{eq:enc_T}
\end{equation}

\vspace{-2mm}

\myparagraph{\textit{Referring Video Object Segmentation}.} RVOS requires the model to not only understand the language description, but also track the referred object in the whole video. To this end, we encode both linguistic and visual information for this task. Similarly, we use Eq.~\ref{eq:enc_T} to extract language feature and apply Eq.~\ref{eq:enc_Vref} and Eq.~\ref{eq:enc_Vmask} for mask features encoding. It should be noted that the mask annotation is available during training. And we use the predicted mask in the previous frame as the visual reference during inference.

\subsection{Multi-scale UniFusion Module} \label{sec:fusion}

After the reference encoding, a natural question is raised: \emph{How to inject the reference information into the network}? In this subsection, 
we introduce our proposed multi-scale UniFusion module for the reference information injection.

The details of UniFusion module is illlustrated in Fig.~\ref{fig:network}\textcolor{red}{b}.
We fuse the visual features $\bm{\mathcal{F}}$ and the reference features in a hierarchical manner. For simplicity, we take the $\ell$-th ($\ell=2,3,4$) visual level for illustration. The UniFusion module receives three inputs: the $\ell$-th level visual feature $\bm{F}_{\ell}$ of current image and the corresponding key and value embeddings ($\bm{F}_{r}^{\rm k}$ and $\bm{F}_{r}^{\rm v}$) from reference features. For mask reference, $\bm{F}_{r}^{\rm k} = \bm{\mathcal{F}}_{V}^{f}$, $\bm{F}_{r}^{\rm v} = \bm{\mathcal{F}}_{V}^{m}$. For language reference, $\bm{F}_{r}^{\rm k} = \bm{F}_{r}^{\rm v} = \bm{F}_{T}$. 
These inputs are first linearly projected and further reformulated as three vectors, namely $\bm{Q}_{\ell}$, $\bm{K}_{\ell}$ and $\bm{V}_{\ell}$. We first perform the mutli-head cross-attention operation between these vectors. Then, reference features $\bm{F}_{r}^{\rm k}$ are pooled and regressed to obtain the scale, shift and gate parameters $\gamma, \beta, \alpha$, which are applied after the attention block. Finally, the output features are injected into the original visual features via residual connection. The process of UniFusion is represented as:

\vspace{-2mm}
\begin{equation}
    \bm{O}_{\ell} = \text{Attention}(\bm{Q}_{\ell}, \bm{K}_{\ell}, \bm{V}_{\ell})
    \label{eq:unifusion_1}
\end{equation}

\vspace{-4mm}
\begin{equation}
    \gamma, \beta, \alpha = \text{Linear(\text{Pooling}($\bm{F}_{r}^{\rm k}$))}
    \label{eq:unifusion_2}
\end{equation}

\vspace{-4mm}
\begin{equation}
    \bm{F}_{\ell}^{\prime} = \bm{F}_{\ell} + \alpha(\bm{O}_{\ell}(1 + \gamma) + \beta)
    \label{eq:unifusion_3}
\end{equation}

\noindent
where $\bm{O}_{\ell}$ is the intermediate results after the attention operation. $\bm{F}_{\ell}^{\prime}$ is the final output of UniFusion. Notably, the UniFusion module shares the same parameters in all visual scales. We emphasize the distinguished characteristics of UniFusion are two-folds: (1) We implement the cross-attention operation using \texttt{FlashAttention}~\cite{dao2022flashattention, dao2023flashattention2}. This leads to the high efficiency and low memory cost when computing on the dense feature maps. (2) Inspired by the adaLN-zero block from \cite{peebles2023dit}, the linear layer for regressing the scale, shift and gate parameters are zero-initialized. This helps network gradually learn the knowledge from references and make UniFusion easily plug in the pretrained object segmentation models.

Thanks to the unifying fusion format, the reference information in different tasks can be injected into the visual features using the same UniFusion module. For the FSS and VOS tasks, the reference features are from the annotated masks. And the reference is language description for RIS. We emphasize here, for RVOS task, both the language features and reference frame visual features are fused with the visual feature of current frame. In this fashion, the network can not only find the referred object by language, but also propagate the target mask across frames for tracking. This also unifies the paradigms of VOS and RVOS to the online pattern.




\begin{table*}[t]
\centering 
\renewcommand\arraystretch{1.10} 
\setlength{\tabcolsep}{2.2mm}    
\small
\caption{Comparison with the state-of-the-art methods on three referring image segmentation (RIS) benchmarks. RN101 denotes ResNet-101~\cite{he2016resnet}, WRN101 refers to Wide ResNet-101~\cite{zagoruyko2016wide}, and DN53 denotes Darknet-53~\cite{redmon2018yolov3}.}
\vspace{-2mm}
\begin{tabular}{l|l|l|l|c|c|c|c|c|c|c|c}
      \toprule[1pt]
      \multicolumn{2}{c|}{\multirow{2}{*}{Method}} & Visual &
      Text & 
      \multicolumn{3}{c|}{RefCOCO}  & \multicolumn{3}{c|}{RefCOCO+} & \multicolumn{2}{c}{RefCOCOg} \\
      \cline{5-12}
                                  \multicolumn{2}{c|}{}  & Backbone & Encoder & val   & test A & test B & val & test A & test B & val-u & test-u  \\
      \hline
      \multirow{13}{*}{\rotatebox{90}{\textbf{oIoU}}} & CMSA~\cite{ye2019cmsa} & RN101 & LSTM & 58.32 & 60.61 & 55.09 & 43.76 & 47.60 & 37.89 & - & - \\
      & STEP~\cite{chen2019see}       & RN101 & Bi-LSTM & 60.04 & 63.46 & 57.97 & 48.19 & 52.33 & 40.41 & -     & -     \\
      & BRINet~\cite{hu2020brinet}    & RN101 & LSTM & 60.98 & 62.99 & 59.21 & 48.17 & 52.32 & 42.11 & -     & -     \\
      & CMPC~\cite{huang2020cmpc}     & RN101 & LSTM & 61.36 & 64.53 & 59.64 & 49.56 & 53.44 & 43.23 & -     & -      \\
      & LSCM~\cite{hui2020lscm} & RN101 & LSTM & 61.47 & 64.99 & 59.55 & 49.34 & 53.12 & 43.50 & -     & -      \\
      & CMPC+~\cite{liu2021cmpc+} &  RN101  & LSTM & 62.47 & 65.08 & 60.82 & 50.25 & 54.04 & 43.47 & -     & -     \\
      & MCN~\cite{luo2020mcn}       & DN53 & Bi-GRU & 62.44 & 64.20 & 59.71 & 50.62 & 54.99 & 44.69 & 49.22 & 49.40    \\
      & EFN~\cite{feng2021efn}                & WRN101 & Bi-GRU & 62.76 & 65.69 & 59.67 & 51.50 & 55.24 & 43.01 & - & -    \\
      & LTS~\cite{jing2021locate}   & DN53 & Bi-GRU & 65.43 & 67.76 & 63.08 & 54.21 & 58.32 & 48.02 & 54.40 & 54.25  \\
      & ReSTR~\cite{kim2022restr}  & ViT-B & Transformer & 67.22 & 69.30 & 64.45 & 55.78 & 60.44 & 48.27 & - & - \\ 
      & LAVT~\cite{yang2022lavt} & Swin-B & BERT-base & 72.73 & 75.82 & 68.79 & 62.14 & 68.38 & 55.10 & 61.24 & 62.09 \\
      \cline{2-12}
      \rowcolor{lightgray!28}& \modelname-R50 & RN50 & RoBERTa-base & 75.04 & 77.28 & 72.43 & 63.25 & 68.12 & 55.56 & 66.96 & 68.77\\
      \rowcolor{lightgray!28}& \modelname-L  & Swin-L & RoBERTa-base & \textbf{79.79} & 81.81 & 77.02 & \textbf{69.26} & \textbf{74.11} & \textbf{63.14} & \textbf{73.04} & \textbf{73.36} \\

      \rowcolor{lightgray!28} & \modelnameplus-R50 & RN50 & BERT-base & 75.63 & 78.75 & 72.91 & 63.29 & 68.68 & 56.26 & 68.38 & 69.71 \\
      \rowcolor{lightgray!28} & \modelnameplus-L & Swin-L & BERT-base & 79.13 & \textbf{82.21} & \textbf{77.45} & 68.43 & 73.98 & 61.45 & 71.37 & 72.84 \\
      
      \hline
      
      \multirow{7}{*}{\rotatebox{90}{\textbf{mIoU}}} & VLT~\cite{ding2021vlt}    &  DN53 & Bi-GRU & 65.65 & 68.29 & 62.73 & 55.50 & 59.20 & 49.36 & 52.99 & 56.65 \\ 
      & CRIS~\cite{wang2022cris}  & RN101 & GPT-2 & 70.47 & 73.18 & 66.10 & 62.27 & 68.06 & 53.68 & 59.87 & 60.36 \\ 
      & SeqTR~\cite{zhu2022seqtr} & DN53 & Bi-GRU & 71.70 & 73.31 & 69.82 & 63.04 & 66.73 & 58.97 & 64.69 & 65.74\\
      & RefTr~\cite{li2021reftr} & RN101 & BERT-base & 74.34 &76.77 & 70.87 & 66.75 & 70.58 & 59.40 & 66.63 & 67.39 \\
      & LAVT~\cite{yang2022lavt}  & Swin-B & BERT-base & {74.46} & {76.89} & {70.94} & {65.81} & {70.97} & {59.23} & {63.34} & {63.62} \\
      & PolyFormer-L~\cite{liu2023polyformer}  & Swin-L & BERT-base  & 76.94 & 78.49 & 74.83 & 72.15 & 75.71 & 66.73 & 71.15 & 71.17 \\
      \cline{2-12}
      \rowcolor{lightgray!28}& \modelname-R50 & RN50 & RoBERTa-base & 78.14 & 80.09 & 75.94 & 69.09 & 73.64 & 62.62 & 71.76 & 73.10\\
      \rowcolor{lightgray!28}& \modelname-L  & Swin-L & RoBERTa-base  & \textbf{81.90} & 83.03 & 79.61 & 73.81 & \textbf{78.30} & \textbf{68.33} & \textbf{76.65} & 77.09 \\

      \rowcolor{lightgray!28} & \modelnameplus-R50 & RN50 & BERT-base & 78.97 & 81.29 & 76.51 & 69.53 & 74.93 & 63.35 & 73.37 & 74.16 \\
      \rowcolor{lightgray!28} & \modelnameplus-L & Swin-L & BERT-base & 81.84 & \textbf{83.48} & \textbf{80.44} & \textbf{74.02} & 78.04 & 68.31 & 76.00 & \textbf{77.20} \\

      \bottomrule[1pt]
\end{tabular}

\label{tab:ris} 
\vspace{-3mm}
\end{table*}

\subsection{Unified Architecture} \label{sec:arch}

The fused multi-scale visual features $\bm{\mathcal{F}}^{\prime} = \left \{  \bm{F}_{\ell}^{\prime} \right \}_{\ell=2}^{4}$ have discriminative representations in highlighting the specific target by reference. We next adopt a unified transform-based architecture to predict the target mask.

\myparagraph{Transformer.} We use the two-stage version Deformable-DETR~\cite{zhu2020deformable-detr} as our object detector. It receives the fused hierarchical visual features $\bm{\mathcal{F}}^{\prime}$ as input and perform multi-scale deformable self-attention in the encoder. In the decoder, $N$ object queries are iteratively refined over stacked decoder layers and converted into the query representations $\bm{Q}_{\text{obj}} \in \mathbb{R}^{N \times C}$ finally. Three predictions heads (class head, box head and mask head) are further built on top of the decoder to predict the object scores $\bm{S} \in \mathbb{R}^{N \times 1}$, boxes $\bm{B} \in \mathbb{R}^{N \times 4}$ and mask dynamic convolution~\cite{tian2020condinst, cheng2021maskformer, cheng2022mask2former} kernel parameters $\bm{\mathcal{G}} = \left \{ \bm{g}_{i} \right \}_{i=1}^{N}$, respectively.

\myparagraph{Mask Decoder.} We take the output features (from strides 8 to 32) of Transformer encoder and hierarchically fuse them in a FPN-like~\cite{lin2017fpn, wu2021seqformer} manner. The feature map with 4$\times$ strides of backbone, namely $\bm{F}_{1}$, is also added in this process. This is helpful for preserving the reference-agnostic and fine-grained information of images. Consequently, we obtain the high-resolution mask features $\bm{F}_{\text{seg}} \in \mathbb{R}^{\frac{H}{4} \times \frac{W}{4} \times C}$. Finally, the masks of target object are generated by performing dynamic convolution between $\bm{F}_{\text{seg}}$ and $\bm{\mathcal{G}}$:

\vspace{-6mm}

\begin{equation}
    \bm{m}_{i} = \text{Upsample}\big(\text{DynamicConv}(\bm{F}_{\text{seg}}, \bm{g}_{i})\big), \; i=1,...,N
    \label{eq:dyanmic_conv}
\end{equation}

During inference, we choose the mask with the highest score as the final result $\bm{m}$ for the target object. Notably, we empirically find that using more object queries leads to higher performance, despite that one object
query is sufficient for the reference-based tasks.

\subsection{Training and Inference} \label{sec:train}

We train \modelnameplus on all related benchmarks of reference-based object segmentation tasks (RIS, FSS, RVOS, VOS). The model with the \emph{same weights} can perform different tasks at run-time by specifying the references.

\myparagraph{Training.} The network predicts $N$ predictions of object scores, box coordinates and segmentation masks, where the object score indicates whether the object is visible in current frame. During training, we apply the set prediction loss~\cite{carion2020detr, zhu2020deformable-detr, sun2021sparse-rcnn} on these predictions. There is only one ground-truth for the reference-based object segmentation tasks. We assign multiple predictions to the ground-truth by selecting the top-$k$ predictions with the least cost according to an optimal transport method~\cite{ge2021ota, ge2021yolox, wu2022idol}. The matching cost is formulated as:

\vspace{-2mm}

\begin{equation}
    \mathcal{C} = \lambda_{cls}  \cdot  \mathcal{C}_{\mathit{cls}} + \lambda_{L1} \cdot \mathcal{C}_{\mathit{L1}} +
    \lambda_{giou} \cdot \mathcal{C}_{\mathit{giou}}
    \label{eq:match}
\end{equation}

\noindent where $\mathcal{C}_{\mathit{cls}}$ is the focal loss~\cite{lin2017focal}. The box losses include the widely-used $\ell_{1}$ loss and generalized IoU (GIoU) loss~\cite{rezatofighi2019giou}. The top-$k$ predictions with the least cost are assigned as positive samples and others as negatives. \modelnameplus is optimized by minimizing the following loss function:

\vspace{-3mm}

\begin{equation}
\begin{aligned}
    \mathcal{L} = \; &\lambda_{cls}  \cdot  \mathcal{L}_{\mathit{cls}} + 
    \lambda_{L1} \cdot \mathcal{L}_{\mathit{L1}} +
    \lambda_{giou} \cdot \mathcal{L}_{\mathit{giou}} \; + \\
    &\lambda_{mask} \cdot \mathcal{L}_{\mathit{mask}} +
    \lambda_{dice} \cdot \mathcal{L}_{\mathit{dice}} 
\end{aligned}
\label{eq:total_loss}
\end{equation}

\vspace{-1mm}

\noindent
where the class loss and boxes losses are the same as those in Eq.~\ref{eq:match}. The mask-related losses contain the mask binary cross-entropy loss and DICE loss~\cite{milletari2016dice}.

\myparagraph{Inference.} For RIS and FSS, we directly output the predicted mask of the query that has the highest score. For RVOS and VOS, our method infers the video in a frame-by-frame online fashion without the complex post-processing. Specifically, for the current frame, the network uses the corresponding references to produce the mask of target object. The mask would be output if its object score is higher than a pre-determined threshold $\sigma$. Otherwise the output mask values are all set to zeros. To handle the videos that contain multiple objects, we adopt the soft-aggregation method commonly used in prior works~\cite{oh2019stm, cheng2021stcn}.
\vspace{-1mm}
\section{Experiments} \label{exps}
\vspace{-1mm}
In this section, we conduct comprehensive experiments on all reference-based tasks (RIS, FSS, RVOS and VOS) to evaluate the effectiveness of our proposed \modelnameplus. The experimental settings will be first introduced in Sec.~\ref{sec:exp_setup}. We then compare \modelnameplus with state-of-the-art methods on the prevalent benchmarks in Sec.~\ref{sec:sota}. The ablation studies are presented in Sec.~\ref{sec:ablation}.

\subsection{Experimental Setup} \label{sec:exp_setup}

\myparagraph{Datasets.} We evaluate our \modelnameplus on four tasks to verify its effectiveness. The specific datasets leveraged in this work for evaluation are presented in the following. \textbf{(i) RIS}: RefCOCO~\cite{yu2016refcoco} consists of 142,209 language descriptions for 50,000 objects in 19,994 images. RefCOCO+~\cite{yu2016refcoco} has 141,564 expressions for 49,856 objects in 19,992 images. RefCOCOg~\cite{mao2016refcocog} includes 85,474 referring expressions for 54,822 objects in 26,711 images. And we use the UMD split for RefCOCOg~\cite{mao2016refcocog}. \textbf{(ii) FSS}: FSS-1000~\cite{li2020fss} is a large-scale dataset for FSS task. It contains 10,000 images from 1,000 classes. \textbf{(iii) RVOS}: Ref-Youtube-VOS~\cite{seo2020urvos} is a large-scale referring video object segmentation dataset which contains 3,978 videos with around 15k langauge descriptions. Ref-DAVIS17~\cite{khoreva2019video} provides the referring expressions for each object in DAVIS17~\cite{pont2017davis}. It contains 90 videos in total. \textbf{(iv) VOS}: Youtube-VOS\footnote{Youtube-VOS and Ref-Youtube-VOS are evaluated using the official server \url{https://youtube-vos.org/}.}~\cite{xu2018youtubevos} is the popular benchmark for video object segmentation. There are 474 and 507 videos in the validation set for 2018 and 2019 version, respectively. LVOS~\cite{hong2022lvos} is a long-term video object segmentation benchmark consisting of 220 videos. The videos in LVOS have an average duration of 1.59 minutes, and the videos in Youtube-VOS last 6 seconds on average. MOSE~\cite{ding2023mose} is a newly proposed dataset for evaluating VOS algorithms in complex scenes, such as occlusion and disappearance. It have 2,149 videos clips and 5,200 objects from 36 categories, with a total of 431,725 annotated masks.

\begin{table}[t]
\centering
\renewcommand\arraystretch{1.10} 
\setlength{\tabcolsep}{3.0mm}    
\small
\caption{Comparison with the state-of-the-art methods on FSS-1000 validation set.}
\vspace{-2mm}

\begin{tabular}{l l | c c}
    \toprule[1pt]
    \multirow{2}{*}{Method} & \multirow{2}{*}{Venue} & \multicolumn{2}{c}{mIoU} \\
     & & 1-shot & 5-shot \\
    \hline
    \multicolumn{2}{l|}{\textit{Specialist Models}} & & \\
    DAN~\cite{wang2020dan} & ECCV'20 & 85.2 & 88.1 \\
    HSNet~\cite{min2021hsnet} & ICCV'21 & 86.5 & 88.5 \\
    SSP~\cite{fan2022ssp} & ECCV'22 & 87.3 & 88.6 \\
    VAT~\cite{hong2022vat} & ECCV'22 & 90.3 & 90.8 \\
    DACM~\cite{xiong2022dacm} & ECCV'22 & 90.8 & 91.7 \\

    \hline
    \multicolumn{2}{l|}{\textit{Generalist Models}} & & \\
    Painter~\cite{wang2023painter} & CVPR'23 & 61.7 & 62.3 \\
    SegGPT~\cite{wang2023seggpt} & ICCV'23 & 85.6 & 89.3 \\
    \rowcolor{lightgray!28} \modelnameplus-R50 & this work & 79.1 & 85.5 \\
    \rowcolor{lightgray!28} \modelnameplus-L & this work & 85.4 & 89.9 \\
    \bottomrule[1pt]

\end{tabular}

\label{tab:fss} 
\vspace{0mm}
\end{table}

\myparagraph{Implementation Details.} We experiment with two prevalent backbones as our visual encoder: ResNet50~\cite{he2016resnet} and Swin Transformer-Large~\cite{liu2021swin}.  The text encoder is selected as BERT-base~\cite{devlin2018bert} and we set the max length of sentences as 77. The Transformer architecture has 6 encoders and 6 decoders with the channel dimension of 256. The number of object queries is set as 300 by default. The loss coefficients in Eq.~\ref{eq:total_loss} are set as $\lambda_{cls}=2.0$, $\lambda_{cls}=2.0$, $\lambda_{L1}=5.0$, $\lambda_{mask}=2.0$ and $\lambda_{dice}=5.0$, respectively.

The entire training process includes three sequential stages, in which pretrained weights from the previous stage are loaded and used for further training. \textbf{(1)} Objects365~\cite{shao2019objects365} pretraining. In this stage, we do not incorporate the UniFusion module but aim to learn a strong object detector for a massive of objects. Due to absence of mask annotation in the dataset, we also apply the BoxInst~\cite{tian2021boxinst} loss for mask supervision. \textbf{(2)} Image-level training. We first combine the training set of RefCOCO/+/g to train the full network, and then train the network for RIS and FSS tasks. \textbf{(3)} Video-level training. At this stage, we randomly sample two frames from a video, where the first frame is considered as the reference frame. To avoid the knowledge forgetting for the RIS task, we also generate pseudo videos for RefCOCO/+/g. The network is jointly trained on all the related benchmarks, including RefCOCO/+/g~\cite{yu2016refcoco, mao2016refcocog}, Ref-YoutubeVOS~\cite{seo2020urvos}, Ref-DAVIS17~\cite{khoreva2019video}, COCO~\cite{lin2014coco}, Youtube-VOS19~\cite{xu2018youtubevos}, OVIS~\cite{qi2022ovis} and LVOS~\cite{hong2022lvos}.

In this work, we use Pytorch toolkit~\cite{paszke2019pytorch} to conduct all experiments on NVIDIA A100 GPUs. Unless otherwise stated, we use $4 \times 8$ A100 GPUs for the objects365 pretraining and $2 \times 8$ GPUs for the following image-level and video-level training. We adopt AdamW~\cite{loshchilov2017adamw} as the optimizer and set the batch size as 2 for each GPU. We refer the readers to Appendix for more implementation details.

\begin{table}[t]
\centering
\renewcommand\arraystretch{1.10} 
\setlength{\tabcolsep}{1.8mm}    
\small
\caption{Comparison with the state-of-the-art methods for referring video object segmentation (RVOS). $^{\dag}$ and $^{\ddag}$ denote the model uses the \texttt{tiny} and \texttt{base} version of Video Swin Transformer~\cite{liu2022video-swin} as visual encoders, respectively.}
\vspace{-2mm}
\begin{tabular}{l|c|ccc}
      \toprule[1pt]
      \multirow{2}{*}{Method} & Visual &
      \multirow{2}{*}{\mjf} & \multirow{2}{*}{\mj} & \multirow{2}{*}{\mj} \\
      & Encoder & & &  \\

     \midrule
     \rowcolor{lightgray!12}
     \multicolumn{5}{c}{\textbf{Ref-Youtube-VOS}} \\
      
      CMSA~\cite{ye2019cmsa} &\multirow{6}{*}{ResNet-50}& 36.4 & 34.8 & 38.1 \\ 
      URVOS~\cite{seo2020urvos} &  & 47.2 & 45.3 & 49.2  \\
      YOFO~\cite{li2022yofo}& & 48.6 & 47.5 & 49.7 \\

    ReferFormer~\cite{wu2022referformer} &  & 58.7 & 57.4 & 60.1 \\
    \rowcolor{lightgray!28}{\modelname-R50} & & 60.6 & 59.0 & 62.3 \\
    \rowcolor{lightgray!28}{\textbf{\modelnameplus-R50}} & & \textbf{61.5} & \textbf{59.7} & \textbf{63.3} \\
    
    \hline 
    
    PMINet + CFBI ~\cite{ding2021pminet} & \multirow{2}{*}{Ensemble} & 54.2 & 53.0 & 55.5\\
    CITD ~\cite{liang2021citd} &  & 61.4 & 60.0 & 62.7  \\
    \hline
    MTTR$^{\dag}$~\cite{botach2022mttr} & \multirow{3}{*}{Video-Swin} & 55.3 & 54.0 & 56.6  \\
    VLT$^{\ddag}$~\cite{ding2022vlt} &  & 63.8 & 61.9 & 65.6  \\
    ReferFormer$^{\ddag}$~\cite{wu2022referformer} &  & 64.9 & 62.8 & 67.0 \\
    \hline
    ReferFormer~\cite{wu2022referformer} &  & 64.2 & 62.3 & 66.2 \\
    \rowcolor{lightgray!28}\textbf{\modelname-L} & Swin-L  & \textbf{67.4} & \textbf{65.5} & \textbf{69.2} \\
    \rowcolor{lightgray!28}\modelnameplus-L & & 66.9 & 64.8 & 69.0 \\
    
    \midrule
    \rowcolor{lightgray!12}
    \multicolumn{5}{c}{\textbf{Ref-DAVIS17}} \\
    
    CMSA~\cite{ye2019cmsa} &\multirow{6}{*}{ResNet-50}& 40.2 & 36.9 & 43.5 \\ 
    URVOS~\cite{seo2020urvos} &  &  51.5 & 47.3 & 56.0 \\
    YOFO~\cite{li2022yofo}& & 54.4 & 50.1 & 58.7\\

    ReferFormer~\cite{wu2022referformer} &  & 58.5 & 55.8 & 61.3 \\
    \rowcolor{lightgray!28} \textbf{\modelname-R50} & & \textbf{63.5} & \textbf{60.0} & \textbf{67.0} \\
    \rowcolor{lightgray!28} \modelnameplus-R50 & & 62.5 & 58.7 & 66.3 \\
    
    \hline
    VLT$^{\ddag}$~\cite{ding2022vlt} & \multirow{2}{*}{Video-Swin} & 61.6 & 58.9 & 64.3 \\
    ReferFormer$^{\ddag}$~\cite{wu2022referformer} &  & 61.1 & 58.1 & 64.1 \\
    \hline
    PolyFormer-L~\cite{liu2023polyformer} & \multirow{3}{*}{Swin-L} & 61.5 & 57.2 & 65.8 \\
    ReferFormer~\cite{wu2022referformer} & & 60.5 & 57.6 & 63.4 \\
    \rowcolor{lightgray!28}\modelname-L &  & 66.3 & 62.9 & 69.7 \\
    \rowcolor{lightgray!28} \textbf{\modelnameplus-L} & & \textbf{67.2} & \textbf{63.4} & \textbf{70.9} \\

     \bottomrule[1pt]
\end{tabular}
\label{tab:rvos} 
\vspace{-4mm}
\end{table}

\begin{table}[t]
\centering
\renewcommand\arraystretch{1.10} 
\setlength{\tabcolsep}{1.8mm}    
\small
\caption{Comparison with the state-of-the-art methods on LVOS~\cite{hong2022lvos} and MOSE~\cite{ding2023mose} validation set. These methods are all not trained on the MOSE dataset.} 
\vspace{-2mm}

\begin{tabular}{l ccc ccc}

	\toprule
	& \multicolumn{3}{c}{LVOS val}  & \multicolumn{3}{c}{MOSE val}  \\
	\cmidrule(lr){2-4} \cmidrule(lr){5-7}
	Method & \mjf & \mj & \mf & \mjf & \mj & \mf\\
	\midrule

        AFB-URR~\cite{liang2020afb-ur} & 34.8 & 31.3 & 38.2 & - & - & - \\
        CFBI~\cite{yang2020cfbi} & 50.0 & 45.0 & 55.1 & - & - & - \\
        RDE~\cite{li2022rde} & 53.7 & 48.3 & 59.2 & 46.8 & 42.4 & 51.3 \\
        STCN~\cite{cheng2021stcn} & 45.8 & 41.1 & 50.5 & 52.5 & 48.5 & 56.6 \\
        AOT~\cite{yang2021aot} & 59.4 & 53.6 & 65.2 & 58.4 & 54.3 & 62.6 \\
        XMem~\cite{cheng2022xmem} & 50.0 & 45.5 & 54.4 & 56.3 & 52.1 & 60.6 \\
        DeAOT~\cite{yang2022deaot} & - & - & - & 59.0 & 54.6 & \textbf{63.4} \\

        \midrule
        \rowcolor{lightgray!28} \modelname-R50 & 55.7 & 51.5 & 60.0 & - & - & - \\
        \rowcolor{lightgray!28} \modelname-L & 60.9 & 57.2 & 64.6 & - & - & - \\
        
        \rowcolor{lightgray!28} \modelnameplus-R50 & 60.1 & 55.8 & 64.3 & 54.7 & 51.3 & 58.2 \\
        \rowcolor{lightgray!28} \textbf{\modelnameplus-L} & \textbf{67.2} & \textbf{62.9} & \textbf{71.5} & \textbf{59.0} & \textbf{55.7} & 62.3 \\
	
	\bottomrule
\end{tabular}

\label{tab:vos_lvos_mose} 
\vspace{0mm}
\end{table}

\subsection{Quantitative Results} \label{sec:sota}

We employed ResNet50~\cite{he2016resnet} and Swin Transformer-Large~\cite{liu2021swin} as visual backbones in our experiments, denoted as \modelnameplus-R50 and \modelnameplus-L, respectively. For each version, all results are computed with one suit of weights. 

\begin{table*}[t]
\centering
\renewcommand\arraystretch{1.10} 
\setlength{\tabcolsep}{2.5mm}    
\small
\caption{Comparison with the state-of-the-art methods on three video object segmentation (VOS) benchmarks.}
\vspace{-2mm}
\begin{tabular}
{l|ccccc|ccccc|ccc}
	\toprule[1pt]
	\multirow{2}{*}{Method} & \multicolumn{5}{c|}{Youtube-VOS 2018 val}  & \multicolumn{5}{c|}{Youtube-VOS 2019 val} & \multicolumn{3}{c}{DAVIS17 val} \\
      \cline{2-14}
	 & \mg & \mjs & \mfs & \mju & \mfu & \mg & \mjs & \mfs & \mju & \mfu & \mjf & \mj & \mf  \\
	
        \midrule \rowcolor{lightgray!8}
        \multicolumn{14}{l}{\textbf{\small{\emph{Memory-based Methods}}}} \\
	STM~\cite{oh2019stm} & 79.4 & 79.7 & 84.2 & 72.8 & 80.9 & -  & -  & -  & -  & -  & 81.8  & 79.2  & 84.3   \\
	AFB-URR~\cite{liang2020afb-ur} & 79.6 & 78.8 & 83.1 & 74.1 & 82.6 & - & -  & -  & -  & - & 76.9 & 74.4 & 79.3 \\
	CFBI~\cite{yang2020cfbi} & 81.4 & 81.1 & 85.8 & 75.3 & 83.4 & 81.0 & 80.6 & 85.1 & 75.2 & 83.0 & 81.9 & 79.1 & 84.6 \\
        RDE~\cite{li2022rde} & - & - & - & - & - & 81.9 & 81.1 & 85.5 & 76.2 & 84.8 & 84.2 & 80.8 & 87.5 \\
	STCN~\cite{cheng2021stcn} & 83.0 & 81.9 & 86.5 & 77.9 & 85.7 & 82.7 & 81.1 & 85.4 & 78.2 & 85.9 & 85.4 & 82.2 & 88.6 \\
        AOT-B~\cite{yang2021aot} & 83.5 & 82.6 & 87.5 & 77.7 & 86.0 & 83.3 & 82.4 & 87.1 & 77.8 & 86.0 & 82.5 & 79.7 & 85.2 \\
        AOT-L~\cite{yang2021aot} & 83.8 & 82.9 & 87.9 & 77.7 & 86.5 & 83.7 & 82.8 & 87.5 & 78.0 & 86.7 & 83.8 & 81.1 & 86.4 \\
	XMem~\cite{cheng2022xmem} & 85.7 & 84.6 & 89.3 & 80.2 & 88.7 & 85.5 & 84.3 & 88.6 & 80.3 & 88.6 & 86.2 & 82.9 & 89.5 \\
        DeAOT~\cite{yang2022deaot} & 86.0 & 84.9 & 89.9 & 80.4 & 88.7 & 85.9 & 84.6 & 89.4 & 80.8 & 88.9 & 85.2 & 82.2 & 88.2 \\
    
     \midrule \rowcolor{lightgray!8}
     \multicolumn{14}{l}{\textbf{\small{\emph{Non-memory Methods}}}} \\
     FRTM~\cite{robinson2020frtm} & 72.1 & 72.3 & 76.2 & 65.9 & 74.1 & - & - & - & - & - & 76.7 & 73.9 & 79.6 \\
     LWL~\cite{bhat2020lwl} & 81.5 & 80.4 & 84.9 & 76.4 & \textbf{84.4} & 81.0 & 79.6 & 83.8 & 76.4 & \textbf{84.2} & 70.6 & 67.9 & 73.3 \\
     \rowcolor{lightgray!28}\modelname-R50 &  81.4 & 81.6 & 85.9 & 75.6 & 82.4 & 81.2 & 80.8 & 84.9 & 76.2 & 83.0 & - & - & - \\
     \rowcolor{lightgray!28}\modelname-L & 82.6 & 83.2 & 87.5 & 76.2 & 83.7 & 82.7 & 82.9 & 86.9 & 76.8 & 84.1 & - & - & - \\

     \rowcolor{lightgray!28} \modelnameplus-R50 & 81.9 & 82.3 & 86.8 & 75.9 & 82.6 & 81.9 & 81.9 & 86.2 & 76.5 & 83.2 & 81.5 & 78.1 & 87.6 \\ 
     \rowcolor{lightgray!28} \textbf{\modelnameplus-L} & \textbf{83.2} & \textbf{83.8} & \textbf{88.5} & \textbf{76.8} & 83.8 & \textbf{83.0} & \textbf{83.1} & \textbf{87.7} & \textbf{77.3} & 84.1 & \textbf{83.9} & \textbf{80.8} & \textbf{89.8} \\ 
     
     \bottomrule[1pt]
\end{tabular}

\label{tab:vos} 
\vspace{0mm}
\end{table*}


\begin{table*}[t]
\renewcommand\arraystretch{0.95} 
\centering
\small
\caption{\textbf{Ablation experiments of \modelname}. We evaluate our model on the RefCOCO, FSS-1000, Ref-Youtube-VOS and Youtube-VOS2018 validation set, respectively. Our default settings are marked in \graybox{gray}.}
\vspace{-5pt}
\label{tab:ablation}
\subfloat[
    Task-specific training.
    \label{tab:ab_task}
]{
    \begin{minipage}{0.30\linewidth}{
    \begin{center}
    \setlength{\tabcolsep}{1mm}
    \begin{tabular}{ccccc}
\toprule[1pt]

\multirow{2}{*}{Training} 
& RIS  & FSS & RVOS &  VOS \\
& oIoU & mIoU & \mjf & \mg \\
\midrule

Signle-task & 72.1  & 81.8 & 58.3 & 81.9 \\
\rowcolor{lightgray!28}Multi-task & 74.7 & 80.3 & 60.1 & 81.9 \\

\bottomrule[1pt]
\end{tabular}
    \end{center}}
    \end{minipage}
}
\subfloat[
    Parameter-sharing for UniFusion.
    \label{tab:ab_fusion}
]{
    \centering
    \begin{minipage}{0.35\linewidth}{
    \setlength{\tabcolsep}{1.0mm}
        \begin{center}
            \begin{tabular}{cccccc}
\toprule[1pt]

\multirow{2}{*}{Tasks} &  \multirow{2}{*}{Levels} 
& RIS & FSS & RVOS &  VOS \\
& & oIoU & mIoU & \mjf & \mg \\
\midrule

\checkmark &  & 73.6 & 64.1 & 60.7 & 81.9 \\
 & \checkmark & 74.9 & 75.2 & 61.5 & 80.9 \\
\rowcolor{lightgray!28}\checkmark & \checkmark &  74.7 & 80.3 & 60.1 & 81.9 \\

\bottomrule[1pt]
\end{tabular}
        \end{center}}
    \end{minipage}
}
\subfloat[
    Query number.
    \label{tab:ab_query}
]{
    \centering
    \begin{minipage}{0.30\linewidth}{
    \setlength{\tabcolsep}{1.0mm}
        \begin{center}
            \begin{tabular}{ccccc}
\toprule[1pt]

\multirow{2}{*}{Query} 
& RIS  & FSS & RVOS &  VOS \\
& oIoU & mIoU & \mjf & \mg \\
\midrule

1 & 72.1 & 75.5 & 58.4 & 79.0 \\
100 & 75.1 & 77.7 & 60.6 & 81.9 \\
\rowcolor{lightgray!28}300 & 74.7 & 80.3 & 60.1 & 81.9 \\

\bottomrule[1pt]
\end{tabular}
        \end{center}}
    \end{minipage}
}
\vspace{-2mm}
\end{table*}

\myparagraph{Referring Image Segmentation.} We compare \modelnameplus with state-of-the-art methods in Table~\ref{tab:ris}. Following the previous works, we use both overall intersection-over-union (oIoU) and mean intersection-over-union (mIoU) as the evaluation metrics. It can be seen that \modelnameplus with ResNet-50 backbone surpasses the previous methods on nearly all splits and it also has significant improvement over UniRef~\cite{wu2023uniref}. When equipped with Swin-Large backbone, \modelnameplus sets new SoTA performance on several splits. For example, our \modelnameplus-L has 4.99 mIoU performance gain over the SoTA method PolyFormer-L~\cite{liu2023polyformer} on the test-A split of RefCOCO. We hypothesize the reason for the similar performance between \modelname and \modelnameplus is that \modelname experienced the Visual Genome pretraining, which makes it more friendly for the grounding tasks.

\myparagraph{Few-shot Segmentation.} 
Following ~\cite{li2020fss, hong2022vat}, we divide the 1,000 classes of FSS-1000~\cite{li2020fss} dataset into 520, 240 and 240 classes, which are used for training, validation and testing, respectively. We evaluate our models on the validation set and report the results in Table~\ref{tab:fss}. It is observed that our models significantly benefit from the few-shot samples and achieve comparable results with the state-of-the-art specialist models.

\myparagraph{Referring Video Object Segmentation.} For the RVOS task, we use the region jaccard \mj\xspace, boundary accuracy \mf\xspace and the average score \mjf\xspace as the evaluation metrics. The comparison of \modelnameplus and state-of-the-art methods are presented in Table~\ref{tab:rvos}. We empirically find that solely using the language as reference would lead to better performance on Ref-DAVIS17, possibly due to salient objects and simple scenes within the dataset. According to Table~\ref{tab:rvos}, it indicates that \modelnameplus has significant improvement over all the previous methods on the two datasets. \modelnameplus with ResNet-50 visual encoder achieves the state-of-the-art performance and has notable 2.8 \mjf\xspace gain over ReferFormer~\cite{wu2022referformer} on Ref-YoutubeVOS. On Ref-DAVIS17, \modelnameplus sets new state-of-the-art performance of 67.2 \mjf\xspace, surpassing the previous best method VLT~\cite{ding2021vlt} by 5.4 \mjf\xspace.

\myparagraph{Video Object Segmentation.} To evaluate the performance on Youtube-VOS~\cite{xu2018youtubevos}, region jaccard \mj\xspace and countour accuracy \mf\xspace are computed for "seen" and "unseen" classes separately, denoted by subscripts $s$ and $u$. \mg\xspace is the average \mjf\xspace for both seen and unseen classes. For DAVIS17~\cite{pont2017davis}, LVOS~\cite{hong2022lvos} and MOSE~\cite{ding2023mose} datasets, \mjf\xspace, \mj\xspace and \mf\xspace are adopted as the evaluation metrics. We provide a comprehensive comparison of different methods on the three classic datasets in Table~\ref{tab:vos}. Our proposed \modelnameplus-L outperforms non-memory-based methods, achieving the best results with 83.2/83.0 \mg\xspace on the Youtube-VOS 2018/2019. Our models also display competitive results on DAVIS17 even if the dataset is not included during training. Unlike memory-based methods~\cite{oh2019stm, cheng2021stcn}, our approach does not rely on predicted masks from past frames, making it more memory-efficient and suitable for long videos. We further demonstrate the advantages of our model for handling long videos in complex scenes in Table~\ref{tab:vos_lvos_mose}. In this situation, \modelnameplus have obvious performance improvement compared with the classic memory-based methods.

\begin{figure*}[t]
\begin{center}
\includegraphics[width=0.98\textwidth]{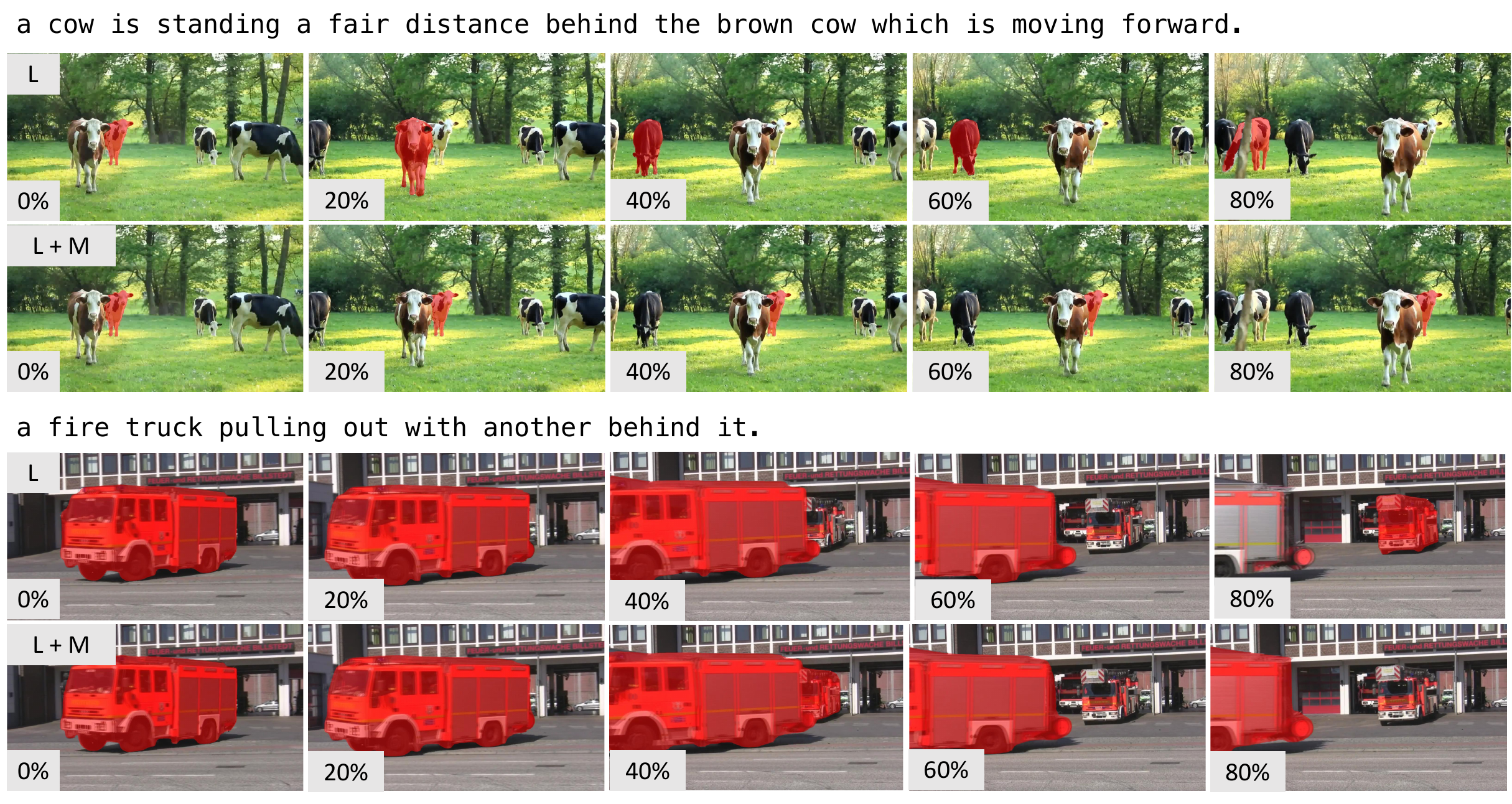}
\end{center}
\vspace{-6mm}
\caption{\textbf{Comparison of the use of mask references for RVOS}. The red masks highlight the predicted objects. The percentages indicate the relative temporal position of each frame in the video. `L' and `M' represent language and mask references, respectively.}
\label{fig:vis-rvos}
\vspace{-3mm}
\end{figure*}

\subsection{Ablation Study} \label{sec:ablation}

The ablation experiments are evaluated on RefCOCO~\cite{yu2016refcoco}, FSS-1000~\cite{li2020fss}, Ref-Youtube-VOS~\cite{seo2020urvos} and Youtube-VOS2018~\cite{xu2018youtubevos} validation set to study \modelnameplus in detail. Unless otherwise stated, we use the ResNet-50 as visual backbone and only conduct the image-level training and video-level training for quick validation. 

\myparagraph{Task-specific Training.} In Table~\ref{tab:ab_task}, we compare results of single training and multi-task joint training. Single-task models are trained on the corresponding datasets, while multi-task models are trained jointly on all datasets. The ablation results indicate that multi-task learning offers significant benefits for both RIS and RVOS. Specifically, Ref-Youtube-VOS achieves 60.1 \mjf\xspace, which is 1.8 points higher than the task-specific model. This can attribute to the fact that the jointly trained model is better at learning mask propagation through VOS training. For FSS, the performance of multi-task model is slightly lower than the single-task model. This is due to that we didn't use the FSS dataset during video-level training. In summary, multi-task joint training improves the performance of task-specific models and saves a significant number of parameters.

\myparagraph{Parameter-sharing for UniFusion Module.} Our \modelnameplus leverages a parameter-sharing UniFusion module to fuse information from both mask and language references for multi-scale visual features. In Table~\ref{tab:ab_fusion}, we present ablation experiments on two weight-sharing variants for UniFusion module. As shown in the table, when we use different UniFusion modules for different tasks, the performance of FSS drops significantly because its data scale is relatively small. When applying different UniFusion for different visual levels, \mg\xspace metric on Youtube-VOS2018 decreases from 81.9 to 80.9. This suggests that a single parameter-sharing UniFusion module is more effective in learning frame similarities for multi-scale visual features.

\myparagraph{Query Number.} Reference-based tasks have one specific target for each reference, making it possible to complete the tasks with just one query. In Table~\ref{tab:ab_query}, we present an ablation study on the number of queries to investigate its impact on performance. As observed, increasing the number of queries from 1 to 100 leads to higher performance. This is reasonable as the model can have more candidates to find the target object, which is particularly helpful in complex scenes where many similar objects co-exist. And we also observe that using 100 and 300 queries would obtain the similar results.

\myparagraph{Does Mask Reference Help RVOS?} For the RVOS task, \modelnameplus processes the videos in an online-fashion. Specifically, we not only use the language reference as guidance, but also leverage the predicted masks in the previous frames for mask propagation. To study the effectiveness of the mask reference, we use our final version models and provide the ablation results in Table~\ref{tab:ab_rvos-vl}. As illustrated in the table, when additionally using the mask references, the model gets 1.1 and 1.4 \mjf\xspace improvement for \modelnameplus-R50 and \modelnameplus-L, respectively. This evidently proves that the mask propagation helps the model to achieve temporal consistency for the target object.

\begin{table}[t]
\centering
\renewcommand\arraystretch{1.10} 
\setlength{\tabcolsep}{1.8mm}    
\small
\caption{\textbf{Ablation on the references used for RVOS}. In the table, `Lang' means language. Results are evaluated on Ref-Youtube-VOS validation set.}
\vspace{-2mm}
\begin{tabular}{c|ccc|ccc}
\toprule[1pt]

\multirow{2}{*}{Reference} & \multicolumn{3}{c|}{ResNet-50}  & \multicolumn{3}{c}{Swin-L} \\
\cmidrule{2-7} 
 & \mjf & \mj & \mf & \mjf & \mj & \mf  \\
\midrule

 Lang & 60.4 & 58.9 & 61.9 & 65.5 & 63.6 & 67.4 \\
 \rowcolor{lightgray!28}Lang + Mask & 61.5 & 59.7 & 63.3 & 66.9 & 64.8 & 69.0 \\

\bottomrule[1pt]
\end{tabular}
\label{tab:ab_rvos-vl} 
\vspace{-4mm}
\end{table}

\begin{table*}[t]
\centering
\renewcommand\arraystretch{1.10} 
\setlength{\tabcolsep}{2.0mm}    
\small
\caption{Comparison with other SAM~\cite{kirillov2023sam}-variant methods. Ref-YT, Ref-D mean Ref-Youtube-VOS and Ref-DAVIS17 datasets. YT-18, YT-19 and D-17 are the abbreations for Youtube-VOS-18, Youtube-VOS-19 and DAVIS17, respectively.} 
\vspace{-2mm}

\begin{tabular}{l| cc| cc| cc| ccc}
    \toprule
    Task & \multicolumn{2}{c|}{RIS} & \multicolumn{2}{c|}{FSS} & \multicolumn{2}{c|}{RVOS} & \multicolumn{3}{c}{VOS} \\
    Dataset & \multicolumn{2}{c|}{RefCOCO val} & \multicolumn{2}{c|}{FSS-1000} & Ref-YT & Ref-D & YT-18 & YT-19 & D-17 \\
    Metric & oIoU & mIoU  & 1-shot & 5-shot & \mjf & \mjf & \mg & \mg & \mjf \\

    \hline
    ReferSAM~\cite{xiao2023refersam} & 64.6 & 71.1 & - & - & - & - & - & - & - \\
    PerSAM~\cite{zhang2023persam} & - & - & 81.6 & - & - & - & - & - & - \\
    PerSAM-F~\cite{zhang2023persam} & - & - & 86.3 & - & - & - & - & - & - \\
    RefSAM~\cite{li2023refsam} & - & - & - & - & 55.1 & 66.1  & - & - & - \\
    SAM-PT~\cite{rajivc2023sam-pt} & - & - & - & - & - & - & 67.5 & - & 76.6 \\

    \hline

    SAM~\cite{kirillov2023sam} + UniRef & 65.9 & 70.4 & 77.8 & 84.2 & 52.4 & 61.2 & 73.4 & 72.8 & 78.4 \\
    \bottomrule

\end{tabular}

\label{tab:sam} 
\vspace{0mm}
\end{table*}

\vspace{-1mm}
\subsection{Qualitative Results}
\vspace{-1mm}

In order to demonstrate the effectiveness of employing mask references in RVOS, we present the qualitative results in Figure~\ref{fig:vis-rvos}. As depicted in the first example, utilizing language references alone struggles in identifying the referred object in a complex scene with multiple similar objects. By integrating mask references, the network can leverage mask propagation to accurately track the target object. This figure illustrates the efficacy of incorporating mask references in RVOS for improving temporal consistency of target object. We also show more visualization examples for VOS and RVOS in the Appendix.

\vspace{-1mm}
\section{Inserting UniFusion into SAM} \label{sec:sam}
\vspace{-1mm}

\begin{figure}[t]
\begin{center}
\includegraphics[width=0.40\textwidth]{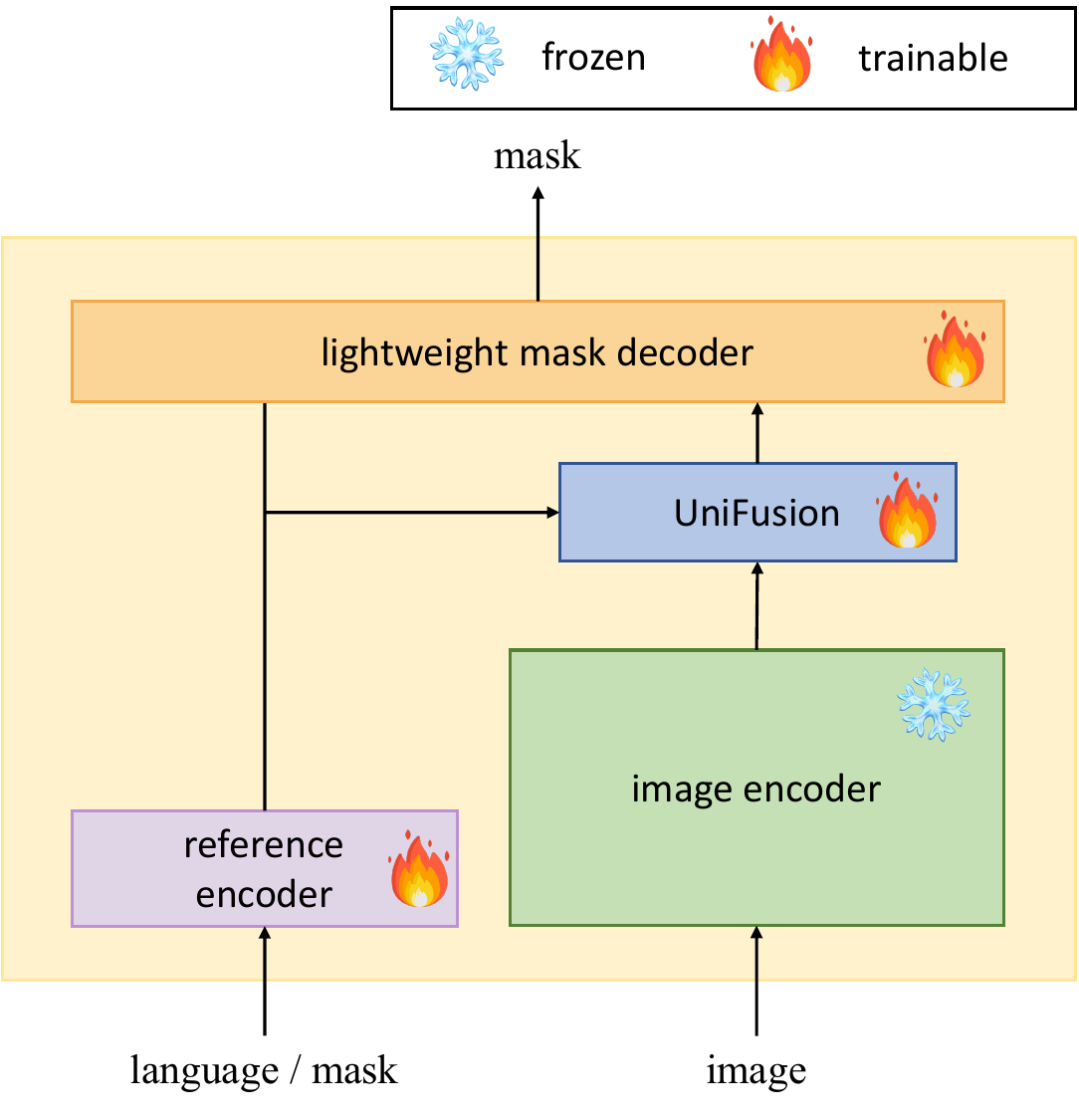}
\end{center}
\vspace{-2mm}
\caption{Our proposed UniFusion module plays as the plug-in component for SAM~\cite{kirillov2023sam}. Analogously, UniFusion could be also easily plugged in other object segmentation foundation models.}
\label{fig:sam}
\vspace{-2mm}
\end{figure}

At the core of \modelnameplus is the UniFusion module for injecting the reference information into the network. To explore its applicability as a plug-in component, we insert the UniFusion module into the advanced foundation model SAM~\cite{kirillov2023sam}. We keep the heavy image encoder frozen and train the lightweight reference encoders, UniFusion and mask decoder. The network goes through the image-level and video-level training as described in Sec.~\ref{sec:exp_setup}. The experiment is highly efficient, being able to be completed within 20 hours using 8 A100s.

We compare the results with other SAM-variant methods in Table~\ref{tab:sam}. It should be noted that other methods are finetuned on the corresponding datasets while our model has one suit of weights. For FSS, our model lags behind the PerSAM~\cite{zhang2023persam} under 1-shot setting since FSS-1000 dataset is not included during video-level training. But the performance gap could be eliminated with few samples (5-shot). From the table, we show that combining SAM~\cite{kirillov2023sam} with UniFusion could achieve satisfactory across the reference-based tasks.

\vspace{-1mm}
\section{Conclusion}
\vspace{-1mm}

We present \modelnameplus, a unified model for four reference-based object segmentation tasks (RIS, FSS, RVOS and VOS). By introducing a UniFusion module to incorporate different types of references, our model can flexibly perform multi-tasks at run-time by specifying the corresponding references and achieves superior performance with a single network. We also show that UniFusion could be play as the plug-in component for the foundation models (\textit{e.g.}, SAM~\cite{kirillov2023sam}) for efficient finetuning.

\section*{Acknowledgements}

This paper is partially supported by the National Key R\&D Program of China No.2022ZD0161000 and the General Research Fund of Hong Kong No.17200622. The paper is supported in part by the National Natural Science Foundation of China under grant No.62293540, 62293542, U1903215 and the Fundamental Research Funds for the Central Universities No.DUT22ZD210.

\clearpage
{\small
\bibliographystyle{ieee_fullname}
\bibliography{egbib}
}

\clearpage
\begin{appendices}

\section{Architecture} \label{sec:app_arch}

\begin{figure}[h]
\begin{center}
\includegraphics[width=0.48\textwidth]{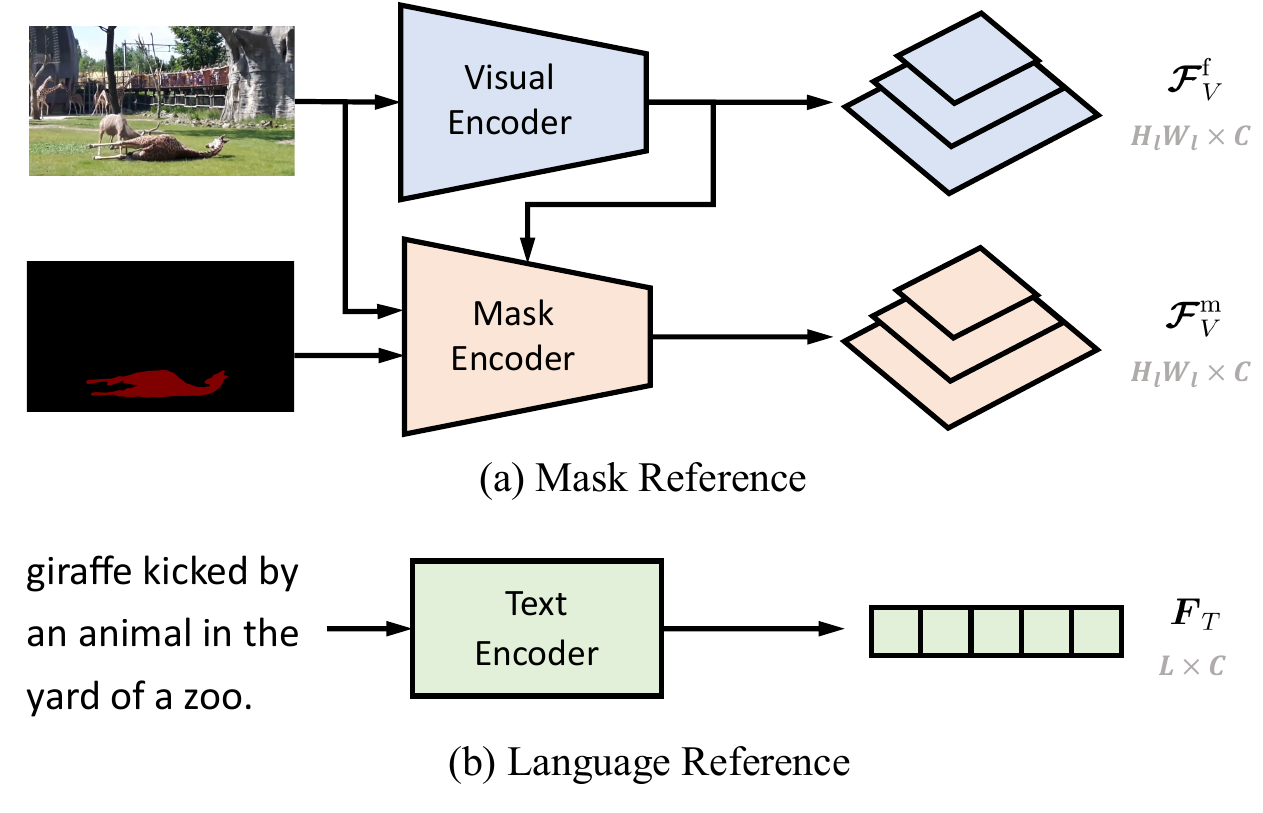}
\end{center}
\vspace{-4mm}
\caption{The process of reference encoding for (a) mask references and (b) language references.}
\label{fig:encoding}
\vspace{-5mm}
\end{figure}

\myparagraph{Reference Encoding} Figure~\ref{fig:encoding} illustrates the process of reference encoding. (i) For mask references, we employ the same visual encoder $\textbf{Enc}_{V}$ for both the current and reference frames to generate multi-scale features (\emph{i.e.}, C3, C4, C5). We denote the encoded features of the reference frame as $\bm{\mathcal{F}_{V}^{\rm f}}$, where the $\ell$-th feature ($\ell=2,3,4$) has a size of $H_{\ell} \times W_{\ell} \times C$, with a spatial stride of $2^{\ell+1}$ relative to the original size. Next, we use a lightweight mask encoder (ResNet-18 in all our experiments) that takes the reference frame and annotated mask as inputs. We concatenate the last three layer features with the corresponding level features in $\bm{\mathcal{F}_{V}^{\rm f}}$ and further process them with two ResBlocks~\cite{he2016resnet} and a CBAM block~\cite{woo2018cbam} to obtain the final outputs, denoted as $\bm{\mathcal{F}_{V}^{\rm m}}$. Finally, we flatten each level feature in $\bm{\mathcal{F}_{V}^{\rm f}}$ and $\bm{\mathcal{F}_{V}^{\rm m}}$ into 1-dimensional vectors. (ii) For language references, we directly use off-the-shelf text encoder BERT~\cite{devlin2018bert} to extract the 1-d linguistic features.

\section{Implementation Details} \label{sec:app_impl} 

\myparagraph{Training Details.} Our training process consists of three sequential stages: Objects365 pretraining, image-level training and video-level training. We train models on NVIDIA A100 GPUs and it takes 5-7 days (depends on the visual backbone) to complete the whole training. The text encoder is unfrozen during the first two stages and then frozen for the final stage. The detailed configurations are summarized in Table~\ref{tab:details}. We follow the implementation of Detic~\cite{zhou2022detic} for the multi-dataset training. The learning rate is reduced by the factor of 10 when the iteration reaches the specified step in the table. Data augmentation includes random horizontal flip and scale jitter for resizing the input images. In the table, short side means the range of values for the shortest side and long side represents the maximum value for the longest side. During video-level training, for COCO~\cite{lin2014coco} and RefCOCO/+/g~\cite{yu2016refcoco, mao2016refcocog}, we apply two different augmentations on the same image to generate the pseudo videos for training. And for OVIS~\cite{qi2022ovis}, we convert the dataset into a class-agnostic format to make it suitable for VOS training.

\myparagraph{Inference Details.} For both RVOS and VOS tasks, all the videos are rescaled to 480p for inference. And the score thresholds are set as 0.4 for VOS datasets and 0.3 for RVOS datasets, respectively. For these two tasks, both the masks in the first frame and previous frame are adopted as references.

\section{More Results} \label{sec:app_results}

\myparagraph{Reference Frames for Mask Propagation.} In this study, we investigate the effect of reference frames for mask propagation, which is presented in Table~\ref{tab:ab_vos}. Specifically, we analyze the impact of discarding the first frame and the previous frame on performance for Youtube-VOS2018 and Ref-Youtube-VOS. These two datasets are evaluated for VOS and RVOS tasks, respectively.
On Youtube-VOS2018, the first frame provides a reliable annotated mask, while the previous frame has the highest similarity with the current frame. Therefore, discarding either of these frames would result in a significant drop in performance. on Ref-Youtube-VOS, there is no ground-truth mask in the first frame. Thus the performance drop is less noticeable. Nevertheless, our findings support the conclusion that utilizing both the first frame and the previous frame as references yields the best results for mask propagation.

\begin{table}[h]
\centering
\renewcommand\arraystretch{1.10} 
\setlength{\tabcolsep}{1.0mm}    
\small
\caption{Ablation on the reference frames used for mask propagation during inference. We use the final model with ResNet-50 visual backbone in this ablation. Our default settings are marked in \graybox{gray}.}
\vspace{-2mm}
\begin{tabular}{cc|ccccc|ccc}
\toprule[1pt]

\multirow{2}{*}{First} & \multirow{2}{*}{Previous} & \multicolumn{5}{c|}{Youtube-VOS2018}  & \multicolumn{3}{c}{Ref-Youtube-VOS} \\
\cmidrule{3-10} 
 & & \mg & \mjs & \mfs & \mju & \mfu & \mjf & \mj & \mf  \\
\midrule

 \checkmark & & 76.6 & 78.7 & 82.8 & 69.6 & 75.2 & 60.8 & 59.1 & 62.6 \\
 & \checkmark & 80.0 & 80.7 & 85.1 & 73.4 & 80.7 & 61.0 & 59.4 & 62.7 \\
 \rowcolor{lightgray!28}\checkmark & \checkmark & 81.9 & 82.3 & 86.8 & 75.9 & 82.6 & 61.5 & 59.7 & 63.3 \\

\bottomrule[1pt]
\end{tabular}
\label{tab:ab_vos} 
\vspace{-2mm}
\end{table}

\begin{table*}[t]
\centering
\renewcommand\arraystretch{1.10} 
\setlength{\tabcolsep}{1.2mm}    
\small
\caption{The detailed configurations for the three training stages.}
\vspace{-2mm}

\begin{tabular}{c|c|ccccc|ccccc}

\toprule[1pt]
Stage & Task & Dataset & 
\multicolumn{1}{c}{\begin{tabular}[c]{@{}c@{}}Sampling\\ Weight\end{tabular}} & 
\multicolumn{1}{c}{\begin{tabular}[c]{@{}c@{}}Batch\\ Size\end{tabular}} & \multicolumn{1}{c}{\begin{tabular}[c]{@{}c@{}}Short\\ Side\end{tabular}}
& \multicolumn{1}{c}{\begin{tabular}[c]{@{}c@{}}Long\\ Side\end{tabular}} & \multicolumn{1}{c}{\begin{tabular}[c]{@{}c@{}}GPU\\ Number\end{tabular}}
& \multicolumn{1}{c}{\begin{tabular}[c]{@{}c@{}}Learning\\ Rate\end{tabular}} 
& \multicolumn{1}{c}{\begin{tabular}[c]{@{}c@{}}Weight\\ Decay\end{tabular}} 
& \multicolumn{1}{c}{\begin{tabular}[c]{@{}c@{}}Max\\ Iteration\end{tabular}} & Step \\

\midrule
\multirow{1}{*} {\uppercase\expandafter{\romannumeral1}}& \multirow{1}{*} {Det} & Objects365~\cite{shao2019objects365} & 1 &  2 & $480\sim800$ & 1333 & 32 & 0.0002 & 0.05 & 340,000 & 310,000 \\
\midrule
\multirow{2}{*} {\uppercase\expandafter{\romannumeral2}} & Det & COCO~\cite{lin2014coco} & 1 & 2 & $480\sim800$ & 1333 & \multirow{3}{*}{16} & \multirow{3}{*}{0.0002} & \multirow{3}{*}{0.05} & \multirow{3}{*}{90,000} & \multirow{3}{*}{75,000} \\
\cline{2-7}
& RIS & RefCOCO/+/g~\cite{yu2016refcoco, mao2016refcocog} & 1 &  2 & $480\sim800$ & 1333 & & & & & \\
\cline{2-7}
& FSS & FSS-1000~\cite{li2020fss} & 0.05 &  2 & $480\sim800$ & 1333 & & & & & \\
\midrule
\multirow{6}{*}{\uppercase\expandafter{\romannumeral3}} & \multirow{4}{*}{VOS} & COCO~\cite{lin2014coco} & 0.40 & 2 & $320\sim640$ & 768 & \multirow{6}{*}{16} & \multirow{6}{*}{0.0001} & \multirow{6}{*}{0.05} & \multirow{6}{*}{90,000} & \multirow{6}{*}{75,000} \\
& & Youtube-VOS2019~\cite{xu2018youtubevos} & 0.30 & 2 & $320\sim640$ & 768 & &&&& \\
& & LVOS~\cite{hong2022lvos} & 0.20 & 2 & $320\sim640$ & 768 & &&&& \\
& & OVIS~\cite{qi2022ovis} & 0.10 & 2 & $320\sim640$ & 768 & &&&& \\
\cline{2-7}
& \multirow{2}{*}{RVOS} & RefCOCO/g/+~\cite{yu2016refcoco, mao2016refcocog} & 0.50 & 2 & $480\sim800$ & 1333   &&&&&\\
& & Ref-Youtube-VOS~\cite{seo2020urvos} & 0.50 & 2 & $320\sim640$ & 768  &&&&&\\

\bottomrule[1pt]
\end{tabular}
    
\label{tab:details} 
\vspace{0mm}
\end{table*}

\begin{figure*}[t]
\centering
\begin{minipage}[t]{0.5\textwidth}
    \vspace{4mm}
    \centering
    \captionsetup{width=0.98\linewidth}
    \captionof{table}{Efficiency comparison of \texttt{FlashAttention}. `YT-VOS18' represents Youtube-VOS2018 dataset. FPS is measured during inference using A100 GPU. Memory is the GPU memory cost during training.}
    \vspace{-2mm}
    \setlength{\tabcolsep}{3.0pt}{
    \scalebox{1.0}{
    \small
    
\begin{tabular}{lcccccc}
    \shline
    \multirow{2}{*}{Dataset} & Mean & Mean & Flash & Num & \multirow{2}{*}{FPS} & \multirow{2}{*}{Memory} \\
    & Frames & Objects & Attention & Heads & & \\
    \hline
    \multirow{4}{*}{YT-VOS18} & \multirow{4}{*}{27} & \multirow{4}{*}{1.9} & \cmark & 1 & 12.4 & 11.6G \\
    & & & \cmark & 8 & 11.1 & 12.7G \\
    & & & \xmark & 1 & 11.9 & 13.4G \\
    & & & \xmark & 8 & 8.3 & 29.9G \\

    \hline
    \multirow{4}{*}{LVOS} & \multirow{4}{*}{574} &  \multirow{4}{*}{1.3}  & \cmark & 1 & 24.4 & 11.6G\\
     & & & \cmark & 8 & 20.5 & 12.7G \\
     & & & \xmark & 1 & 23.4 & 13.4G \\
     & & & \xmark & 8 & 15.9 & 29.9G \\

    \shline
\end{tabular}
}}
    \label{tab:fps}
\end{minipage}
\begin{minipage}[t]{0.48\textwidth}
    \vspace{6mm}
    \centering
    \resizebox{!}{0.58\linewidth}{
	\begin{tikzpicture}
	\begin{axis}[
		xlabel={Number of processed frames},
		ylabel={FPS},
		xmin=25,xmax=7000,
		ymin=0,
		ymode=log,
		ytick={0.1, 1, 10, 20, 30, 100},
		minor ytick={2,3,4,5,6,7,8,9,40,50,60,70,80,90},
		yticklabels={0.1, 1, 10, 20, 30, 100},
		grid=both,
		height=8cm, width=12cm,
		legend style={font=\footnotesize,at={(0.7,0.4)},anchor=west},
		legend cell align={left},
		]

		\addplot[each nth point={3},mark=o,color=red,mark size=1.5pt]
		coordinates {
			(50),(19.6)
(100),(19.6)
(150),(19.6)
(200),(19.6)
(250),(19.6)
(300),(19.6)
(350),(19.6)
(400),(19.6)
(450),(19.6)
(500),(19.6)
(550),(19.6)
(600),(19.6)
(650),(19.6)
(700),(19.6)
(750),(19.6)
(800),(19.6)
(850),(19.6)
(900),(19.6)
(950),(19.6)
(1000),(19.6)
(1050),(19.6)
(1100),(19.6)
(1150),(19.6)
(1200),(19.6)
(1250),(19.6)
(1300),(19.6)
(1350),(19.6)
(1400),(19.6)
(1450),(19.6)
(1500),(19.6)
(1550),(19.6)
(1600),(19.6)
(1650),(19.6)
(1700),(19.6)
(1750),(19.6)
(1800),(19.6)
(1850),(19.6)
(1900),(19.6)
(1950),(19.6)
(2000),(19.6)
(2050),(19.6)
(2100),(19.6)
(2150),(19.6)
(2200),(19.6)
(2250),(19.6)
(2300),(19.6)
(2350),(19.6)
(2400),(19.6)
(2450),(19.6)
(2500),(19.6)
(2550),(19.6)
(2600),(19.6)
(2650),(19.6)
(2700),(19.6)
(2750),(19.6)
(2800),(19.6)
(2850),(19.6)
(2900),(19.6)
(2950),(19.6)
(3000),(19.6)
(3050),(19.6)
(3100),(19.6)
(3150),(19.6)
(3200),(19.6)
(3250),(19.6)
(3300),(19.6)
(3350),(19.6)
(3400),(19.6)
(3450),(19.6)
(3500),(19.6)
(3550),(19.6)
(3600),(19.6)
(3650),(19.6)
(3700),(19.6)
(3750),(19.6)
(3800),(19.6)
(3850),(19.6)
(3900),(19.6)
(3950),(19.6)
(4000),(19.6)
(4050),(19.6)
(4100),(19.6)
(4150),(19.6)
(4200),(19.6)
(4250),(19.6)
(4300),(19.6)
(4350),(19.6)
(4400),(19.6)
(4450),(19.6)
(4500),(19.6)
(4550),(19.6)
(4600),(19.6)
(4650),(19.6)
(4700),(19.6)
(4750),(19.6)
(4800),(19.6)
(4850),(19.6)
(4900),(19.6)
(4950),(19.6)
(5000),(19.6)
(5050),(19.6)
(5100),(19.6)
(5150),(19.6)
(5200),(19.6)
(5250),(19.6)
(5300),(19.6)
(5350),(19.6)
(5400),(19.6)
(5450),(19.6)
(5500),(19.6)
(5550),(19.6)
(5600),(19.6)
(5650),(19.6)
(5700),(19.6)
(5750),(19.6)
(5800),(19.6)
(5850),(19.6)
(5900),(19.6)
(5950),(19.6)
(6000),(19.6)
(6050),(19.6)
(6100),(19.6)
(6150),(19.6)
(6200),(19.6)
(6250),(19.6)
(6300),(19.6)
(6350),(19.6)
(6400),(19.6)
(6450),(19.6)
(6500),(19.6)
(6550),(19.6)
(6600),(19.6)
(6650),(19.6)
(6700),(19.6)
(6750),(19.6)
(6800),(19.6)
(6850),(19.6)
(6900),(19.6)
(6950),(19.6)
(7000),(19.6)
(7050),(19.6)
(7100),(19.6)
(7150),(19.6)
(7200),(19.6)
		};
		\addlegendentry{UniRef}

		\addplot[each nth point={5},mark=*,mark=diamond*,color=blue]
		coordinates {
(25),(69.91420212241066)
(50),(31.42355310045728)
(75),(16.151106097297625)
(100),(10.938407719024825)
(125),(7.9142476247090485)
(150),(6.60727002853302)
(175),(5.593538957228665)
(200),(4.809135640587997)
(225),(4.132455684644876)
(250),(3.756250515618619)
(275),(3.2559115973996455)
(300),(3.0755787241810904)
(325),(2.8707750681468776)
(350),(2.638893841786615)
(375),(2.41846472502204)
(400),(2.2052814017986178)
(425),(2.08352553465083)
(450),(1.9911651087174256)
(475),(1.8379268299779565)
(500),(1.8116675442438137)
(525),(1.7179796634131128)
(550),(1.608350632266829)
(575),(1.5599185943666278)
(600),(1.4660824629160802)
(625),(1.4436967214185739)
(650),(1.3823966439081476)
(675),(1.2905879412486208)
(700),(1.2810356497682234)
(725),(1.225377511038408)
(750),(1.183619331363938)
(775),(1.1342169322712483)
(800),(1.0825579039778688)
(825),(1.0600459837566627)
(850),(1.0159643766018687)
(875),(0.9858499743434334)
(900),(0.9773067460295936)
(925),(0.9290347649204285)
(950),(0.897977350071046)
(975),(0.8896353271125931)
(1000),(0.8689599314283266)
(1025),(0.8313886183052289)
(1050),(0.8104567858624226)
(1075),(0.8084107106586204)
(1100),(0.8070124910552978)
(1125),(0.7766282373292159)
(1150),(0.7563999529962341)
(1175),(0.7407031974408069)
(1200),(0.7170081648352932)
(1225),(0.7034080403177011)
(1250),(0.6830143458702793)
(1275),(0.6800641594190928)
(1300),(0.66451443811702)
(1325),(0.6516491796735416)
(1350),(0.6471616110992324)
(1375),(0.6341593244987825)
(1400),(0.6243431846892161)
(1425),(0.5967941913646901)
(1450),(0.5887093186430284)
(1475),(0.5754725546763029)
(1500),(0.5626369067749135)
(1525),(0.5565409535225387)
(1550),(0.5470298383122562)
(1575),(0.5350537441538644)
(1600),(0.5278984213598087)
(1625),(0.5127403866030711)
(1650),(0.5093794739167136)
(1675),(0.5078151469489693)
(1700),(0.49974186164547696)
(1725),(0.49919407304537106)
(1750),(0.4824019037820066)
(1775),(0.47925626208617483)
(1800),(0.48016451144217104)
(1825),(0.47097749668680006)
(1850),(0.46295014096086)
(1875),(0.45867384828301794)
(1900),(0.4548046193835364)
(1925),(0.44718304673977094)
(1950),(0.43768783732913713)
(1975),(0.4303325480521651)
(2000),(0.4251918895665667)
(2025),(0.41749170545713643)
(2050),(0.4179040372645979)
(2075),(0.4091413796829202)
(2100),(0.4065244393540668)
(2125),(0.398803346550104)
(2150),(0.3963797880223814)
(2175),(0.38865810822863056)
(2200),(0.38800750510870047)
(2225),(0.3810497797586521)
(2250),(0.3839907170994123)
(2275),(0.3713188432428771)
(2300),(0.3768445942641042)
(2325),(0.3683877746759796)
(2350),(0.3645696689374008)
(2375),(0.35581415037126846)
(2400),(0.35113736047788835)
(2425),(0.32236905841411845)
		};
		\addlegendentry{STCN}
		
	\addplot+[no markers,color=blue,domain=10:7000]{1/(0.0012*x - 0.0329)};
		
	\end{axis}
\end{tikzpicture}
    }
    \vspace{-3ex}
    \captionsetup{width=0.9\linewidth}
    \caption{FPS scaling of our method and the representative memory-based method STCN.}
    \label{fig:fps-scaling}
\end{minipage}
\vspace{-4mm}
\end{figure*}

\myparagraph{Efficiency Comparison.} We compare the efficiency of using \texttt{FlashAttention}~\cite{dao2022flashattention, dao2023flashattention2} on two VOS datasets, namely Youtube-VOS18~\cite{xu2018youtubevos} and LVOS~\cite{hong2022lvos}, as displayed in Table \ref{tab:fps}. The results clearly show that \texttt{FlashAttention} could improve the FPS during inference, especially when employing the multi-head attention. Also, it can greatly reduce the GPU memory cost during training, consuming only 12.7G with 8 heads.

Our method has a constant memory cost, while the classic memory-based methods have linear memory complexity with respect to the video duration. This suggests that our method is more efficient for the long-term videos. To better highlight the advantages of our method, we further compare our method with the representative memory-based method STCN in terms of the single-object FPS-scaling in Figure \ref{fig:fps-scaling}.


\section{Visualization Results} \label{sec:app_vis}

\begin{figure*}[t]
\begin{center}
\includegraphics[width=0.98\textwidth]{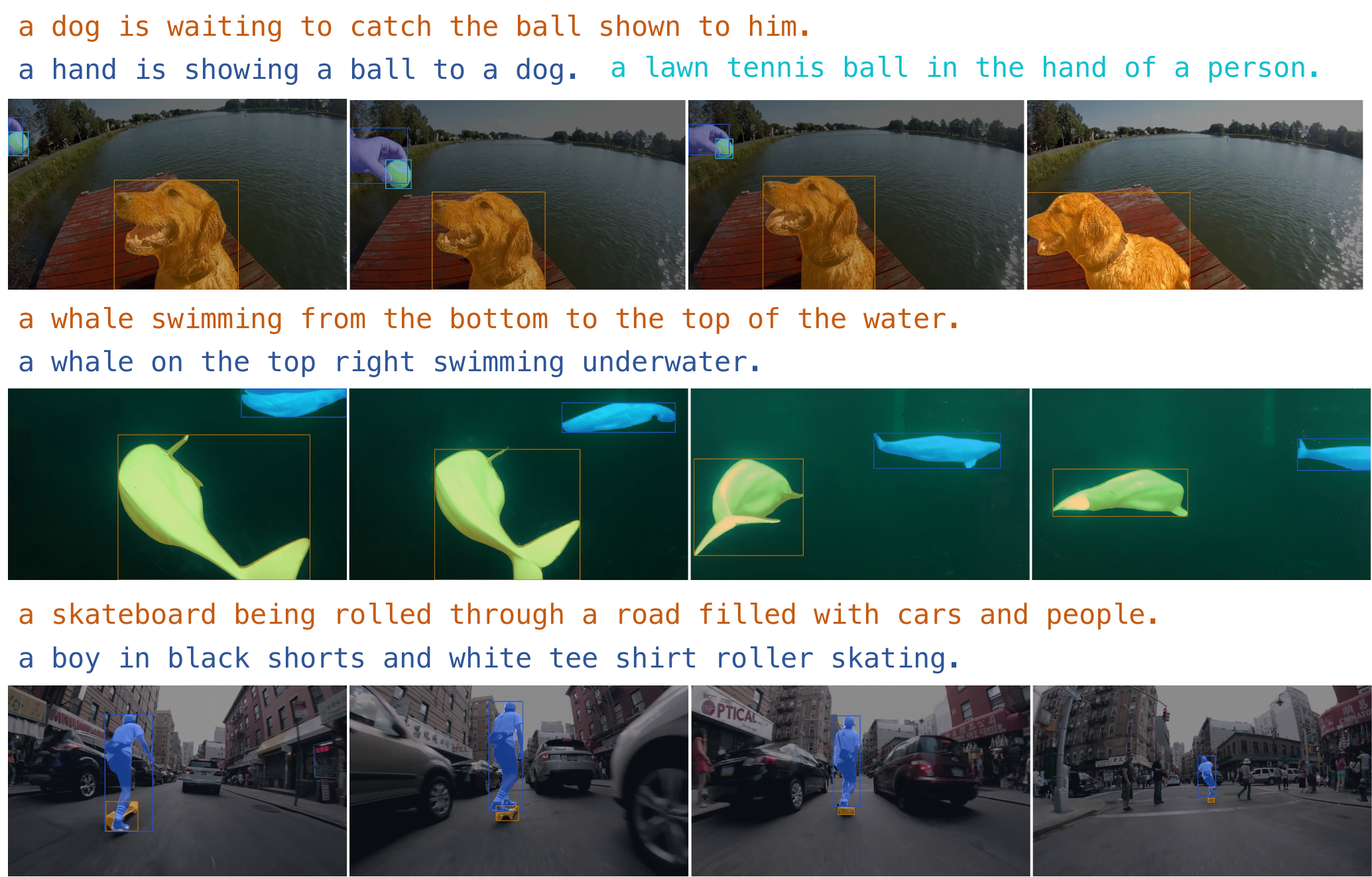}
\end{center}
\vspace{-4mm}
\caption{Visualization results on Ref-Youtube-VOS validation set.}
\label{fig:rvos-refytvos}
\vspace{-2mm}
\end{figure*}

\begin{figure*}[t]
\begin{center}
\includegraphics[width=0.98\textwidth]{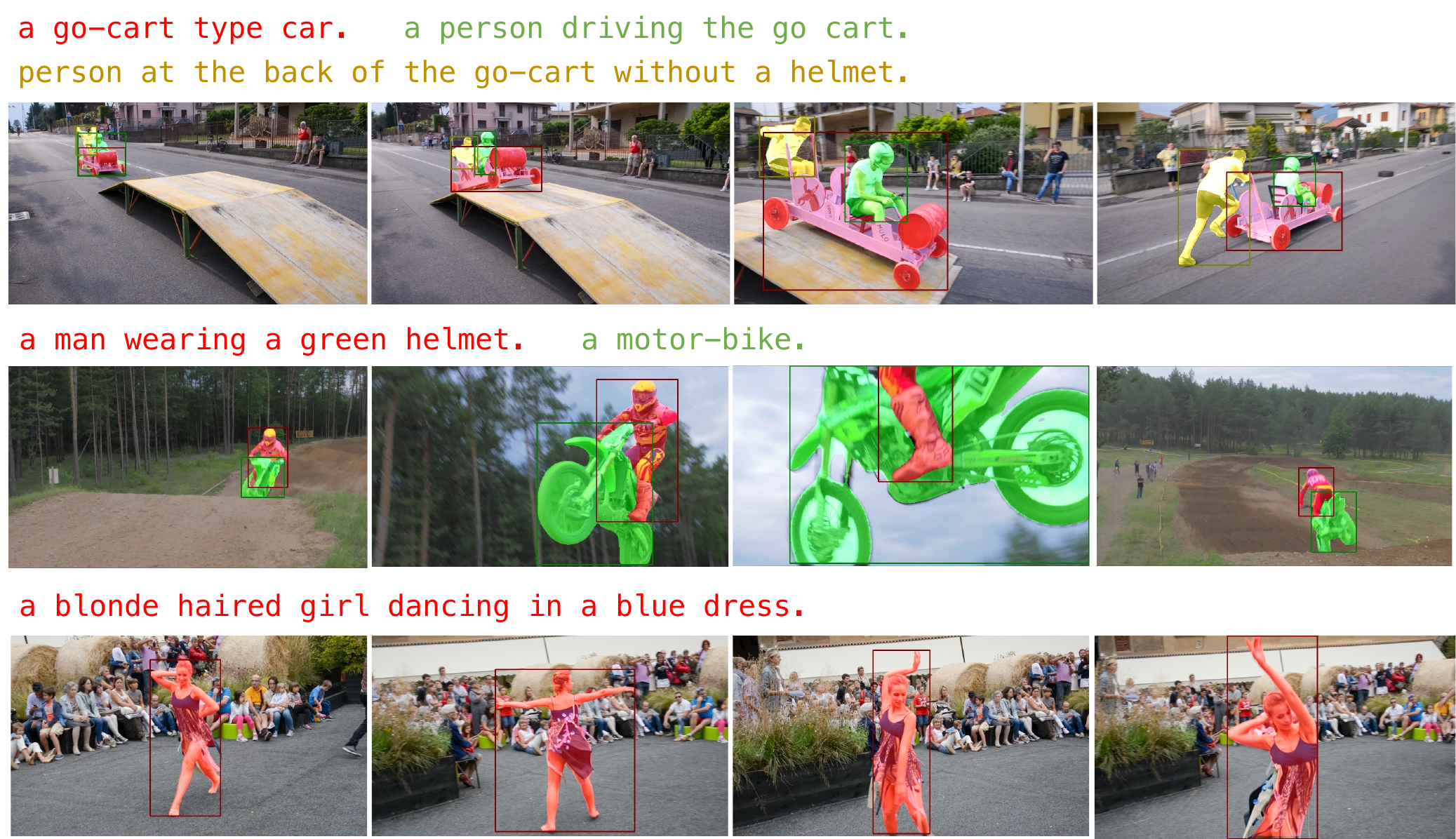}
\end{center}
\vspace{-4mm}
\caption{Visualization results on Ref-DAVIS17 validation set.}
\label{fig:rvos-refdavis}
\vspace{-5mm}
\end{figure*}

We provide the visualization results of \modelname-L for RVOS tasks in Figure~\ref{fig:rvos-refytvos} and Figure~\ref{fig:rvos-refdavis}. It can be seen that our model can segment the referred objects correctly and accurately in various challenging scenes, \emph{e.g.}, partial display, similar objects and fast moving, as illustrated in Figure~\ref{fig:rvos-refytvos}.

\begin{figure*}[t]
\begin{center}
\includegraphics[width=0.98\textwidth]{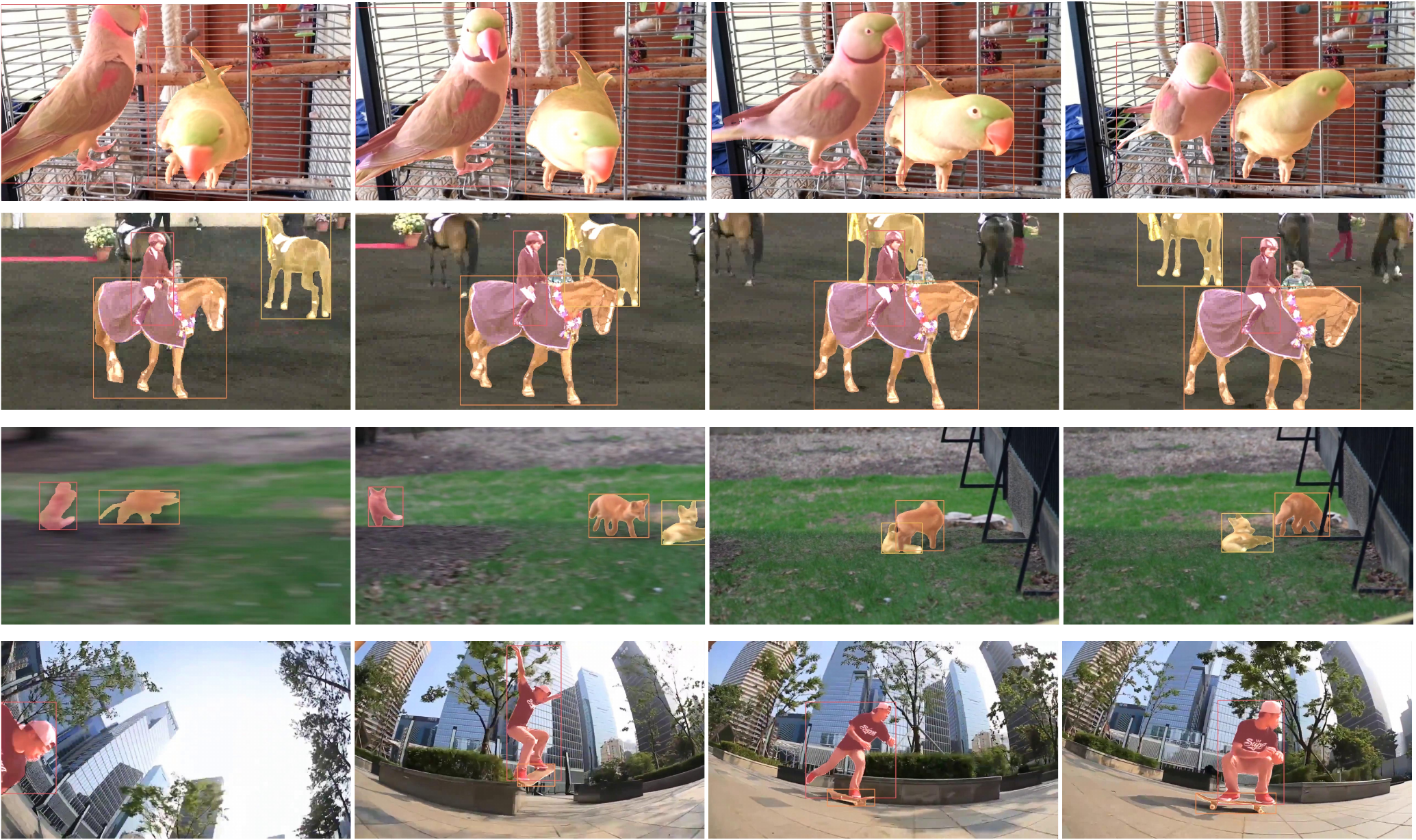}
\end{center}
\vspace{-4mm}
\caption{Visualization results on Youtube-VOS2018 validation set.}
\label{fig:vos-ytbvos18}
\vspace{0mm}
\end{figure*}

\begin{figure*}[t]
\begin{center}
\includegraphics[width=0.98\textwidth]{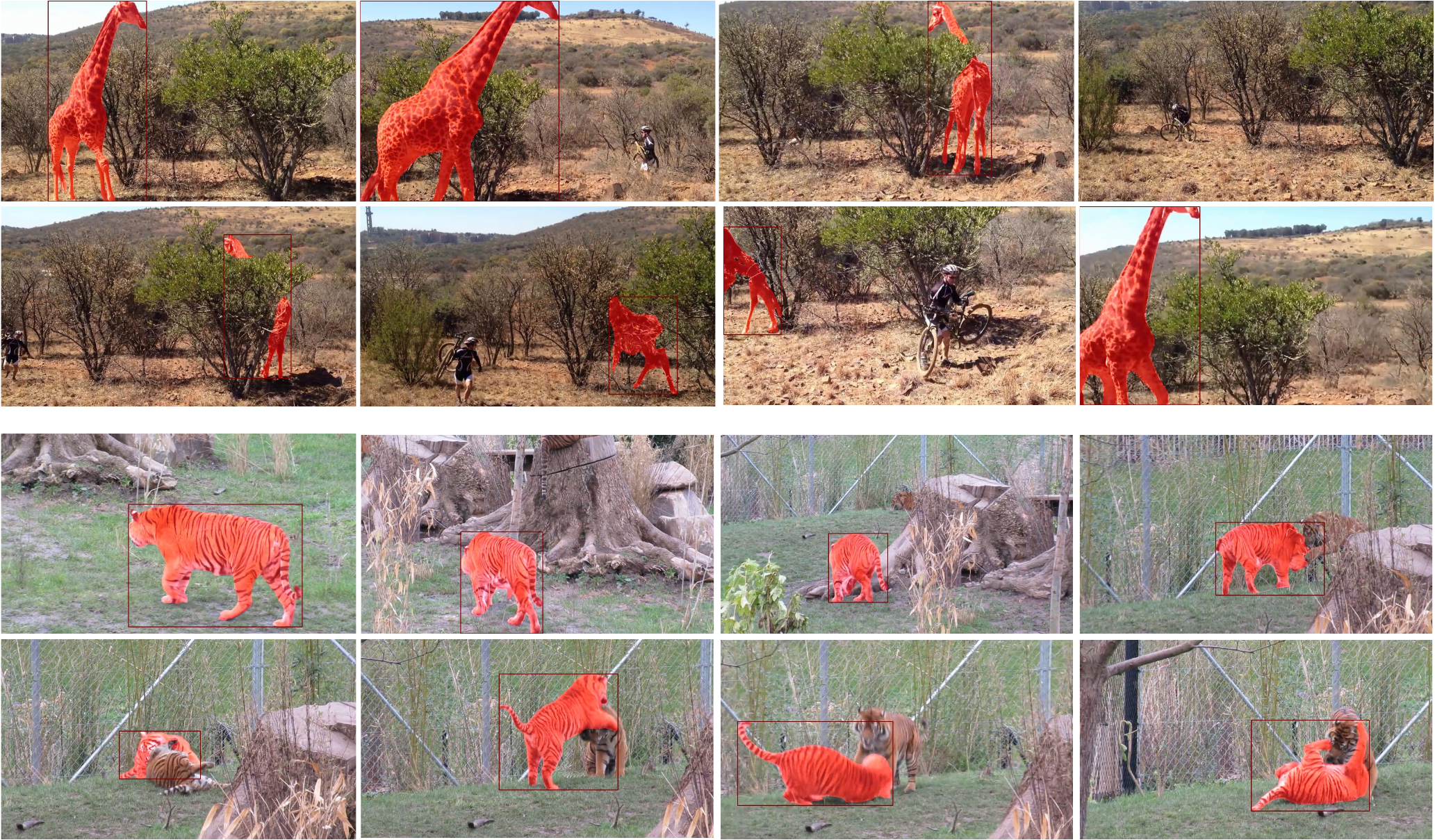}
\end{center}
\vspace{-4mm}
\caption{Visualization results on LVOS validation set.}
\label{fig:vos-lvos}
\vspace{-5mm}
\end{figure*}

Visualization results for the VOS tasks are presented in Figure~\ref{fig:vos-ytbvos18} and Figure~\ref{fig:vos-lvos}. Notably, our model reveals strong ability in handling long-term videos that typically last for over a minute, such as those in LVOS~\cite{hong2022lvos}. As shown in Figure~\ref{fig:vos-lvos}, our model can accurately segment the target objects throughout the whole video, despite the objects have significant pose variation. We further provide a video demo in the supplementary material.



\end{appendices}

\end{document}